%% file: icml_tmp.tex
\documentclass{article}

\usepackage{microtype}
\usepackage{graphicx}
\usepackage{subfigure}
\usepackage{booktabs} 
\usepackage{enumitem}

\usepackage{hyperref}

\usepackage[accepted]{icml2021}

\icmltitlerunning{Online KSD Thinning}

\usepackage[ruled,vlined]{algorithm2e}
\usepackage{algorithmic}
\usepackage{amsmath}
\usepackage{amsmath}

\DeclareMathOperator*{\argmin}{arg\,min}
\usepackage{amsopn}
\usepackage{amssymb}
\usepackage{xcolor}

\newtheorem{lemma}{Lemma}
\newtheorem{theorem}{Theorem}
\newtheorem{corollary}{Corollary}
\newtheorem{definition}{Definition}
\newcommand{\mat}[1]{\mathbf{#1}}

\newcommand{\dict}{\mat{D}}
\newcommand{\seq}{\mat{S}}
\newcommand{\ksd}{{\rm KSD}}
\usepackage{pifont}
\newcommand{\cmark}{\ding{51}}%
\newcommand{\xmark}{\ding{55}}%

\begin{document}

\twocolumn[
\icmltitle{Online, Informative MCMC Thinning with Kernelized Stein Discrepancy}

\icmlsetsymbol{equal}{*}

\begin{icmlauthorlist}
\icmlauthor{Cole Hawkins}{ucsb}
\icmlauthor{Alec Koppel}{amazon}
\icmlauthor{Zheng Zhang}{ucsb}
\end{icmlauthorlist}

\icmlaffiliation{ucsb}{University of California, Santa Barbara}
\icmlaffiliation{amazon}{Amazon Supply Chain Optimization Technologies}
\icmlcorrespondingauthor{Cole Hawkins}{colehawkins@math.ucsb.edu}

\icmlkeywords{Machine Learning, ICML}

\vskip 0.3in
]

\begin{abstract}
A fundamental challenge in Bayesian inference is efficient representation of a target distribution. Many non-parametric approaches do so by sampling a large number of points using variants of Markov Chain Monte Carlo (MCMC). We propose an MCMC variant that retains only those posterior samples which exceed a KSD threshold, which we call KSD Thinning. We establish the convergence and complexity tradeoffs for several settings of KSD Thinning as a function of the KSD threshold parameter, sample size, and other problem parameters. Finally, we provide experimental comparisons against other online nonparametric Bayesian methods that generate low-complexity posterior representations, and observe superior consistency/complexity tradeoffs. Our attached code will be made public.
\end{abstract}

\input{sections/background}

\input{sections/algorithm}

\input{sections/convergence_main}

\input{sections/experiments}

\input{sections/conclusion}

\newpage
\bibliography{references}
\bibliographystyle{icml2021}

\onecolumn
\input{sections/convergence_appendix}
\newpage
\input{sections/experiments_appendix}
\newpage
\input{sections/misc_appendix}

\end{document}

%% file: sections/background.tex
\section{Introduction}\label{sec:intro}

Uncertainty quantification aids automated decision-making by permitting risk evaluation and expert deferral in applications such as medical imaging and autonomous driving. Nonparametric Bayesian inference methods such as Markov Chain Monte Carlo (MCMC) are the gold standard in uncertainty estimation problems, but sample complexity is a major bottleneck in their practical application. A major limitation of MCMC is that the samples generated by a transition kernel are correlated, which can lead to redundancy in the constructed estimate. 

In uncertainty quantification each retained sample corresponds to one expensive forward simulation \citep{constantine2016accelerating,peherstorfer2018survey,martin2012stochastic} and in machine learning each retained sample requires the storage and inference costs of a expensive model such as a neural network \citep{neal2012bayesian,lakshminarayanan2016simple}. Therefore balancing {\it representation quality} and {\it representational complexity} is an important tradeoff. In standard MCMC, to ensure statistical consistency, the representational complexity approaches infinity. Classically, to deal with the redundancy issue, one may employ online or post-hoc ``thinning," which discards all but a random subset of MCMC samples \cite{raftery1996implementing,link2012thinning}. More recently this task is sometimes called ``quantizing" a posterior distribution~\citep{riabiz2020optimal}, especially when the work studies Bayesian cubature~\citep{teymur2020optimal}. Existing approaches generate a large set of samples (usually via MCMC) and then ``thin" the sample set. Doing so may require storing a large number of samples before the final post-processing stage~\citep{riabiz2020optimal,teymur2020optimal} and does not allow the sampler to target a compressed representation during the MCMC iterations. Traditional online thinning is done without any goodness-of-fit metric on the samples, which may ignore gradient information generated by modern stochastic gradient MCMC methods \citep{welling2011bayesian,ma2015complete}. 


To select samples to retain during thinning, one may compute metrics between the empirical measure and the unknown target $\mathbb{P}$. However, in a Bayesian inference context, many popular integral probability metrics (IPM) are not computable. The \emph{kernelized Stein discrepancy} (KSD) addresses this by employing the score function of the target combined with reproducing kernel Hilbert Space (RKHS) distributional embeddings to define statistics that track the discrepancy between distributions in a computationally feasible manner \citep{berlinet2011reproducing,sriperumbudur2010hilbert,stein1972bound,liu2016kernelized}. 
\subsection{Related Work}
Several MCMC thinning procedures have been developed based upon RKHS embedding: Offline methods such as Stein Thinning and MMD thinning \citep{riabiz2020optimal,teymur2020optimal,dwivedi2021kernel} take a full chain $\seq$ of MCMC samples as input and iteratively build a subset $\dict$ by greedy KSD/MMD minimization. The online method Stein Point MCMC (SPMCMC) \citep{chen2019stein} selects the optimal sample from a batch of $m$ samples during MCMC sampling: at each step it adds the best of $m$ points to a $\dict$. Doing so mitigates both the aforementioned redundancy and representational complexity issues; however \cite{chen2019stein} only append new points to the existing empirical measure estimates, which may still retain too many redundant points. Another online approach maintains a fixed-size reservoir of samples, but the reservoir size is pre-fixed \cite{paige2016super}. 

A similar line of research in Gaussian Processes \citep{williams2006gaussian} and kernel regression \citep{hofmann2008kernel} reduces the complexity of a nonparametric distributional representation through offline point selection rules such as Nystr\"{o}m sampling \citep{NIPS2000_19de10ad}, greedy forward selection \citep{seeger2003fast,JMLR:v13:wang12b}, or inducing inputs \citep{snelson2005sparse}. In Gaussian processes fixing the approximation error instead of the representational complexity, and allowing both constructive/destructive operations in the dictionary selection, yields improved distributional estimates when employing online sample processing \citep{koppel2017parsimonious,koppel2020consistent}.

Most similar to our work is a variant of Stein Point MCMC proposed in \citep{chen2019stein}[Appendix A.6.5] which develops a non-adaptive add/drop criterion where a pre-fixed number of points are dropped at each stage and the dictionary size grows linearly with the time step. By contrast, our work develops a flexible and adaptive scheme that automatically determines the number of points to drop and the dictionary size grows sub-linearly with the time step.

\subsection{Contributions}


\begin{table*}
\small
\begin{center}
\begin{tabular}{ c| c| c | c | c}
 Method & Online & Informative  &  Discard Past Samples & Model Order Growth \\ \hline
 MCMC Thinning & \cmark & \xmark & \xmark & $n$ \\  \hline
 Stein Thinning~\citep{riabiz2020optimal} & \xmark & \cmark & \cmark & NA \\  \hline
 SPMCMC~\citep{chen2019stein}         & \cmark & \cmark & \xmark & $n$   \\  \hline
This work & \cmark & \cmark & \cmark & $o\left(\sqrt{n\log(n)}\right)$  \\ \hline 
\end{tabular}
\end{center}
\caption{Methods for Generating Compressed Non-Parametric Representations\label{tab: comparison}}
\end{table*}

\begin{figure}[t]
     \centering
    \includegraphics[width=0.5\linewidth]{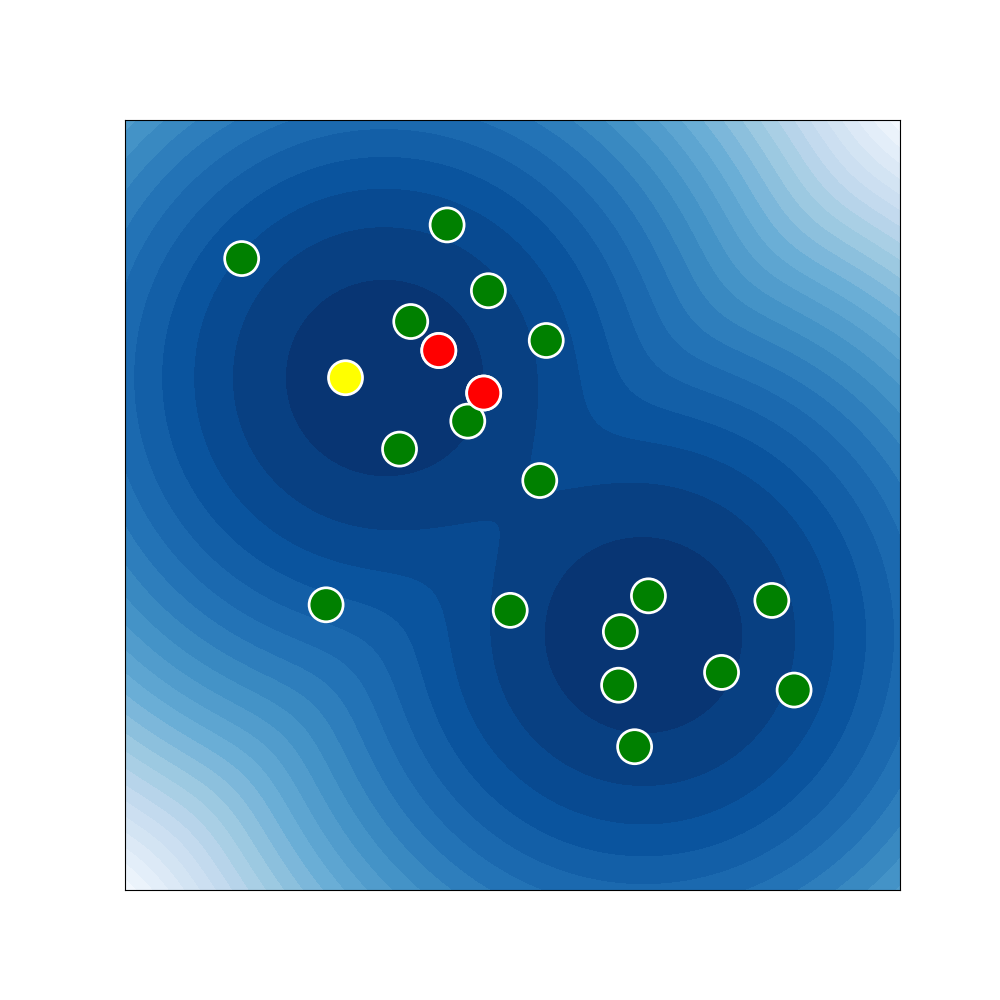}
        \caption{Current dictionary $\dict_{t-1}$ in green. New sample $\mat{x}_t$ added in yellow to form $\tilde{\dict}_t$. Then redundant red samples are thinned.} 
        \label{fig: alg depiction}
\end{figure}

We propose a method that generates a compressed representation by {\bf flexible and online} thinning of a stream of MCMC samples. 
Our method specifies a KSD budget parameter which determines both the transient dictionary complexity and asymptotic bias of the inference by allowing both constructive and destructive point selection during sampling, rather than after the fact. In contrast to existing online approaches that require linear growth in the size of the active set we require only  $o\left(\sqrt{n\log(n)}\right)$ growth to ensure convergence through a novel memory-reduction routine we call Kernelized Stein Discrepancy Thinning (KSDT). We compare our approach and several others in Table \ref{tab: comparison}. Our method, described at a high level in Figure \ref{fig: alg depiction}, enables sample-efficient Bayesian learning by directly targeting compressed representations during the sampling process. We make the following specific contributions: 
\begin{itemize}
    \item [\bf C1.] We introduce the first online thinning algorithm that can provide informative removal of past MCMC samples during the sampling process. Our algorithm permits a flexible tradeoff between model order growth, thinning budget, and posterior consistency.
    \item [\bf C2.] We prove in Theorem \ref{thm: decaying budget} that our thinning method can be applied to existing SOTA MCMC algorithms with no change in asymptotic convergence rate when the thinning budget asymptotically decays to $0$. In Corollary \ref{thm: constant budget convergence corollary} we provide the KSD neighborhood of convergence when the thinning budget is fixed.
    \item [\bf C3.] We test our method on two MCMC problems from the biological sciences and two Bayesian Neural Network problems and demonstrate that our thinning algorithm can reduce the number of retained samples but retain or improve baseline sampler performance. 
\end{itemize}

%% file: sections/algorithm.tex

\section{Online KSD Thinning}\label{sec:algorithm}



Given a sequence of points $\seq=\{\mat{x}_i\}_{i=1}^N$ drawn from an unknown probability measure $\mathbb{P}$ with $\mat{x}_i\in\mathcal{X}\subset\mathbb{R}^d$, our goal is to infer  an approximate empirical measure $q_\dict=\hat{\mathbb{P}}$. Here $q_\dict$ is a particle representation with a sparse \emph{dictionary} $\dict\subsetneq\seq$, i.e., $|\dict|\ll N$, where $N\leq \infty$ is the potentially infinite sample size. 
Specifically, the approximate density associated with $\dict$ takes the form
\begin{equation}
    \label{eq: approximating distribution}
    q_\dict(\cdot) = \frac{1}{|\dict|}\sum_{\mat{x}_i\in \dict} \delta_{\mat{x}_i}(\cdot),
\end{equation}
where $\delta_{\mat{x}_i}$ denotes the Dirac delta which is $1$ if its argument is equal to $\mat{x}_i$, and $0$ otherwise. 

We assume that the measure $\mathbb{P}$ admits a density $p$ which can only be evaluated up to an unknown normalization constant. That is, $p=\tilde{p}/Z$ with $\tilde{p}$ as the unnormalized density and $Z>0$ as the normalization constant. We assume that the unnormalized density $\tilde{p}$ and its score function $\nabla \log \tilde{p}$ may be evaluated in a computationally affordable manner. This set of assumptions is standard in many Bayesian inference problems that arise in machine learning and uncertainty quantification \cite{welling2011bayesian,chen2019stein,constantine2016accelerating,peherstorfer2018survey}. 

Our goal is construct $\dict$ from the stream of samples $\seq$ by determining which points to retain from $\seq$ and discard from $\dict$. The empirical distribution $q_\dict$ can then be used to approximate integrals of the form 
\begin{equation}
    \label{eq: evaluation goal}
    \int_\mathcal{X} f(\mat{x})p(\mat{x})d\mat{x}\approx \frac{1}{|\dict|}\sum_{\mat{x}_i\in\dict} f(\mathbf{x}_i).
\end{equation} 

\subsection{Kernelized Stein Discrepancy}


Informative thinning requires a goodness-of-fit metric to distinguish between ``good'' samples and ``bad'' samples. In general, evaluation of metrics between the empirical estimate $q_\dict$ and the target $p$ is intractable. The powerful combination of Stein's method \cite{stein1972bound} and Reproducing Kernel Hilbert Spaces \cite{berlinet2011reproducing,sriperumbudur2010hilbert} permits one to evaluate discrepancies between $q_\dict$ and $p$ in closed form \cite{liu2016kernelized}. Let $k$ be a base kernel, for example the inverse multi-quadratic (IMQ) kernel 
\begin{equation}
    \label{eq: imq kernel}
    k(\mat{x},\mat{y})=\left(1+\|\mat{x}-\mat{y}\|_2^2\right)^{-0.5}
\end{equation} 
or the Radial Basis Function (RBF) kernel
\begin{equation}
    \label{eq: rbf kernel}
    k(\mat{x},\mat{y})=\exp\left(-\|\mat{x}-\mat{y}\|_2^2/2h\right).
\end{equation} 
Next define the {\em Stein kernel}
\begin{equation}
    \label{eq: stein kernel}
    \begin{split}
    k_0(\mat{x},\mat{y}) = &\nabla_{\mat{x}} \log p(\mat{x})^T\nabla_{\mat{y}}\log     p(\mat{y})k(\mat{x},\mat{y})\\
    &+\nabla_{\mat{y}}\log p(\mat{y})^T\nabla_\mat{x} k(\mat{x},\mat{y})\\
        &+\nabla_{\mat{x}}\log p(\mat{x})^T\nabla_{\mat{y}} k(\mat{x},\mat{y})
        +\sum_{i=1}^d \frac{\partial^2 k(\mat{x},\mat{y})}{\partial \mat{x}_i\partial\mat{y}_i}\; ,
    \end{split} 
\end{equation}
where $p$ is a fixed target density, $d$ is the dimension of each particle $\mat{x}, \mat{y}\in \mathcal{X}$ and the score function $\nabla_\mat{x} \log p(\mat{x})$ can be estimated without knowledge of the normalizing constants $Z$. Stein's method in this context specifies a constructed RKHS $\mathcal{K}_0$ with Stein kernel $k_0$ \eqref{eq: stein kernel} in turn constructed from the base kernel $k$. Then the kernelized Stein discrepancy (KSD) of an empirical measure $q_\dict$ with respect to a target density $p$ is the RKHS norm in $\mathcal{K}_0$ given by
\begin{equation}
    \label{eq: ksd} 
        \ksd(q_\dict)=\sqrt{\frac{1}{n^2}\sum_{\mat{x}_i,\mat{x}_j\in \dict} k_0(\mat{x}_i,\mat{x}_j)}.
\end{equation}
For simplicity our notation suppresses the dependence on the target density $p$ and the RKHS $\mathcal{K}$. 

\subsection{Online Thinning - Outer Loop}

\begin{algorithm}[t]
\caption{Online KSD Thinning (KSDT)}
\begin{algorithmic}
\label{alg: outer loop}
\REQUIRE Target density $p$, initial dictionary $\dict_0$, point sequence $\seq=\{\mat{x}_i\}_{i=1}^\infty$, budget sequence $\{\epsilon_t\}_{t=1}^\infty$, minimum sample size function $f$.
\FOR{$t$ in $[1,2,3,\dots]$}
\STATE Receive new sample $\mat{x}_t$ from $\seq$
\STATE Add sample $\mat{x}_t$ to dictionary: $\tilde\dict_t = \dict_{t-1}\cup \{\mat{x}_t\}$
\STATE Thin via Algorithm \ref{alg: inner loop}: $\dict_t=\texttt{ KSDT}(p,\tilde\dict_t,\epsilon_t,f(t))$
\ENDFOR
\end{algorithmic}
\end{algorithm}
\begin{figure}[t]
    \centering
    \includegraphics[width=0.3\textwidth]{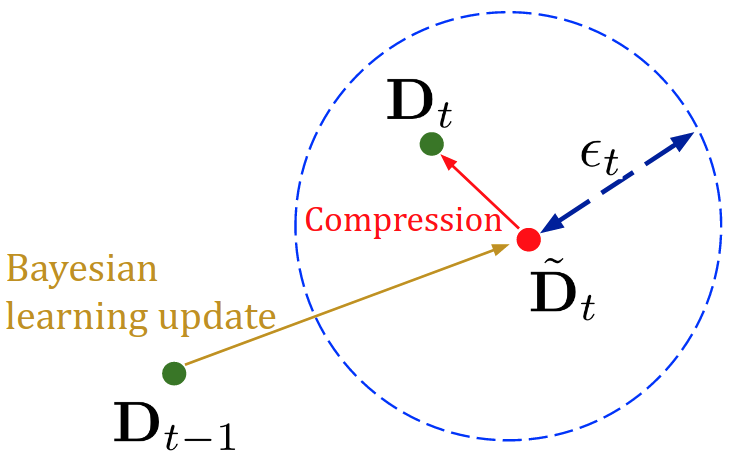}
    \caption{A geometric view of online KSD thinning (Algorithm \ref{alg: outer loop}). We update $\dict_{t-1}$, fix a KSD $\epsilon$-neighborhood, and greedily remove points until we hit the boundary of this neighborhood. By tuning the compression-induced error to the information contained in the update direction, i.e., the red arrow representing Eq. \eqref{eq: add sample update}, we obtain the green dot $\dict_t$, which is the update output by Algorithm \ref{alg: inner loop}.} 
    \label{fig: high level alg}
\end{figure}

In Algorithm \ref{alg: outer loop} we propose an online thinning algorithm that can generate compressed representations of a measure $\mathbb{P}$ with access only to a stream of samples $\seq$, the unnormalized density $\tilde p$, and the score function $\nabla\log\tilde p$. Our sampler performs informative thinning in a flexible online fashion by discarding points that do not make a sufficient contribution to minimizing the KSD objective $\ksd(q_{\dict_t})$ where $\dict_t$ is the thinned dictionary produced by step $t$ of our algorithm.

At each time step $t$ we take previous dictionary $\dict_{t-1}$ and add a new sample $\mat{x}_t$ from our MCMC chain, which results in the expanded auxiliary dictionary:
\begin{equation}
    \label{eq: add sample update}
    \tilde\dict_t=\dict_{t-1}\cup\{\mat{x}_t\} \; , 
\end{equation}
which we compress via a destructive inner loop. This outer loop pseudo-code is given in Algorithm \ref{alg: outer loop}.
If the sample stream $\seq$ is generated by an MCMC method and we skip the destructive thinning inner loop then $\dict_t=\tilde\dict_t$ and the update rule exactly matches standard MCMC.

At each step the thinning budget schedule $\epsilon_t$ controls the compression/fidelity tradeoff of the thinned approximation. The minimum dictionary size function $f(t)$ must satisfy the relationship $f(t)= \Omega\left(\sqrt{t\log(t)}\right)$ to preserve the consistency our algorithm. However, if one tolerates non-vanishing asymptotic bias, then $f(t)$ may be set to a small constant $\epsilon_t=\epsilon$. We discuss practical selection of $\epsilon$ and the resultant dictionary $\dict$ in the following section.
Our approach differs from the related work of \cite{chen2019stein} in that we thin the entire dictionary at each step. In contrast, no prior samples are thinned in the Stein Point MCMC algorithm presented by \cite{chen2019stein}. By thinning the entire dictionary we can remove previously sampled points that are inessential for ensuring the representation is consistent. A major advantage of this approach is that our online thinning algorithm can automatically determine the number of points required based on the complexity of the target measure $\mathbb{P}$. We demonstrate this in Section \ref{sec: adaptivity}.

\subsection{Online Thinning - Inner Loop}
\begin{algorithm}[t]
\caption{Destructive KSD Thinning}
\begin{algorithmic}
\label{alg: inner loop}
\REQUIRE Target density $p$, empirical measure dictionary $\dict$, budget $\epsilon$, minimum sample number $S$
\STATE Compute reference KSD: $M=\ksd(q_\dict)$
\WHILE{$\ksd(q_\dict)^2<M^2+\epsilon$ and $|\dict|>S$}
\STATE Compute least influential point $\mat{x}_j$ as in \eqref{eq: least influential} 
\IF{$\ksd(q_{\dict \setminus \{\mat{x}_j\}})^2<M^2+\epsilon$}
\STATE Remove least influential point, set $\dict = \dict\setminus\{\mat{x}_j\}$
\ELSE
\STATE Break loop
\ENDIF
\ENDWHILE \\
\Return thinned dictionary $\dict$ satisfying $\ksd(q_\dict)^2<M^2+\epsilon$
\end{algorithmic}
\end{algorithm}

The inner loop performs destructive thinning on the auxiliary dictionary $\tilde\dict_t$ based on a maximum KSD thinning budget $\epsilon_t$ and a minimum dictionary size $f(t)$.

Given an intermediate dictionary $\tilde\dict_t$ our goal is to return a compressed representation $\dict_t\subset{\tilde\dict_t}$ that satisfies 
\begin{equation}
    \label{eq: ksd goal}
    \ksd(q_{\dict_t})^2< \ksd(q_{\tilde\dict_t})^2+\epsilon.
\end{equation}
The compressed dictionary $\dict_t$ must be $\epsilon$-close in squared KSD to the uncompressed intermediate dictionary $\tilde\dict_t$. In this section we determine how to build $\dict_t$.

Related work by \cite{riabiz2020optimal,teymur2020optimal} present {\it constructive} greedy approaches to subset selection during post-hoc thinning. The challenge with online constructive approaches is time complexity, as the inner loop requires $|\tilde\dict_{t}|$ point selections in the worst case (no points are thinned). We expect that few points will be thinned at each step and so we follow a destructive approach that requires only $|\dict_t|-|\tilde\dict_{t}|$ point selections per step. In practice, only 0-2 points are thinned at each step and so $|\dict_t|-|\tilde\dict_{t}|\in \{0,1,2\}$ while $|\dict_t|$ ranges from 10-1000. 

Our destructive approach iteratively removes points by selecting the ``least influential" point in $\dict_t$
\begin{equation}
    \label{eq: least influential}
\mat{x}_j=\text{argmin}_{\mat{x}\in \dict_t} \ksd(q_{\dict_t\setminus\{\mat{x}\}})
\end{equation}
If removing the least influential point would violate the KSD criterion in \eqref{eq: ksd goal} we do not thin the point and we break the thinning loop. If removing the least influential point does not violate the KSD criterion we update $\dict_t\leftarrow \dict_t\setminus \{\mat{x}_j\}$ and repeat the thinning procedure. The inner loop pseudo-code is given in Algorithm \ref{alg: inner loop} and visualized in Figure \ref{fig: alg depiction}. 

Thinning a single point in this manner every $k$ steps was suggested by \cite{chen2019stein} in Appendix A.6.5 of their work. The main difference between their work and ours is flexibility: they thin a fixed number of points and do not use a KSD-aware thinning criterion (our budget parameter $\epsilon_t$) and so ``good'' points may be unnecessarily removed or too few ``bad" points may be removed. Their work does not present convergence guarantees for this thinning strategy. The worst-case per-step time complexity of our method is $\mathcal{O}\left(t\right)$ which matches \cite{chen2019stein} (see Appendix \ref{sec: computational complexity}).

\subsection{Sample Stream Filtering Requirement}

In addition to our practical goal of informative online thinning, we aim to provide an algorithm with provable convergence. We will thin samples, potentially incurring a KSD increase $\epsilon_t$ at each step, we require that our sample stream $\seq$ already converges in KSD. Therefore we use the SPMCMC approach of \cite{chen2019stein} to generate a sample stream $\seq$ that provably and rapidly converges in KSD.  

{\bf SPMCMC Update Rule}: Let $\dict_{t-1}$ be the current dictionary and $\mat{y}_{t,0}$ be the current point. To construct $\tilde\dict_{t}$ we initialize an MCMC chain at $\mat{y}_{t,0}$, generate $m$ MCMC samples $\mathcal{Y}_t=\{\mat{y}_{t,l}\}_{l=1}^m$, and append the KSD-optimal point in $\mathcal{Y}_t$ to $\dict_{t-1}$:  
\begin{equation}
    \label{eq: SPMCMC update rule}
    \begin{split}
        \mat{x}_t&=\argmin_{\mat{y}\in \mathcal{Y}_t}\ksd(\dict_{t-1}\cup\{\mat{y}\}).\\
        \tilde\dict_{t}&=\dict_{t-1}\cup\{\mat{x}_t\}.
    \end{split}
\end{equation}
In Section \ref{sec: experiments body} we provide empirical evidence that our method is suitable for non-SPMCMC samplers (eg. MALA, RWM).

%% file: sections/convergence_main.tex
\section{Convergence}\label{sec: convergence body}
\begin{table*}
\small
\begin{center}
\begin{tabular}{ c| c| c | c }
Method & Per-step bound  & Per-step dictionary size  & Thinning \\ \hline
SPMCMC~\citep{chen2019stein} & $\log(n)/n$ & $n$ &\xmark \\  \hline
Ours & $n\log(n)/f(n)^2$ & $f(n)= \Omega\left(\sqrt{n\log(n)}\right)$ & \checkmark \\ 
\end{tabular}
\end{center}
\caption{Quantitive Comparison with SPMCMC\label{tab: spmcmc comparison}. Our proposed approach offers a complexity/consistency tradeoff based on the model order growth $f$ and enables online thinning. When $f(n)\propto n$ our approach achieves the SPMCMC convergence rate but enables thinning.}
\end{table*}

This section provides convergence guarantees for our proposed KSD-Thinning method (Algorithm \ref{alg: outer loop}) under several settings. We consider two sample generation mechanisms: i.i.d. sampling and MCMC sampling, and two settings for the sequence of budget parameters $\{\epsilon_t\}$: decaying budget ($\epsilon_t\rightarrow 0)$ and fixed budget ($\epsilon_t=\epsilon$). The decaying budget setting is most applicable when one desires exact posterior consistency in infinite time, whereas fixed budgets are useful when a specified limiting error tolerance is sufficient. In both decaying and fixed budget settings we prove that:
\vspace{-0.05in}
\begin{itemize}
    \setlength\itemsep{0em}
    \item Our KSD-Thinning algorithm maintains the same asymptotic KSD convergence rate of the baseline sampler, but gains the ability to thin past samples.
    \item Our algorithm permits a complexity/consistency trade-off that enables higher particle reduction in exchange for the cost of a slower asymptotic convergence rate.
\end{itemize} 
\vspace{-0.05in}
The convergence guarantees and complexity/consistency trade-offs of the two thinning budget settings are provided in Theorem \ref{thm: decaying budget} and Corollary \ref{thm: decaying budget}, respectively. The most relevant comparison to our work is \cite{chen2019stein} which achieves an $\mathcal{O}(\log(n)/n)$ rate of KSD convergence. Our main result in Theorem \ref{thm: decaying budget} differs from the results of \cite{chen2019stein} in two respects. If we maintain the same asymptotic convergence rate, we can also thin previous samples. Second, our analysis enables the user to trade off between sublinear $\sqrt{n\log(n)}$ growth of $|\dict_t|$ instead of the linear growth in $|\dict_t|$ required by \cite{chen2019stein} at the cost of a slower asymptotic convergence rate. The main technical novelty of our proof is the decomposition of our per-step bound into two terms: the sampling error term from \citep{chen2019stein} and a thinning error term introduced by Algorithm \ref{alg: inner loop}. We defer proofs to Appendix \ref{sec: convergence appendix}.


\subsection{Preliminaries}
\label{sec: convergence preliminaries}

Before we present our main result we introduce two new definitions to impose standard requirements from \cite{gorham2017measuring,chen2019stein} on the Metropolis-Adjusted Langevin Algorithm (MALA) sampler and the distribution $p$. Informally, we require that that target density does not possess a heavy tail (distantly dissipative) and that accepted proposals have low norm (inwardly convergent). These conditions hold for the MALA sampler targeting standard smooth densities.

\begin{definition}
\label{def: distantly dissipative}
A density $p$ with lipschitz score function $\nabla \log \tilde p$ is {\bf distantly dissipative} if 
\begin{equation}
    \label{eq: distantly dissipative}
    \begin{split}
    \liminf_{r\rightarrow 0} \inf_{\|\mat{x}-\mat{y}\|\leq r}\left\{\frac{\langle \nabla \log \tilde p(\mat{x})-\nabla \log \tilde p(\mat{y}),\mat{x}-\mat{y}\rangle}{\|\mat{x}-\mat{y}\|_2^2}\right\}>0.\\
    \end{split}
\end{equation}
\end{definition}
Any density that is strongly log-concave outside of a compact set is distantly dissipative, and a common example is a Gaussian mixture \citep{korba2020non}.

\begin{definition}
\label{def: inwardly convergent}
Let $t(\mat{x},\mat{y})$ denote the MALA transition kernel, and $\alpha(\mat{x},\mat{y})$ denote the probabilty of accepting proposal $\mat{y}$ given current state $\mat{x}$. Let $\mathcal{A}(\mat{x})=\{\mat{y}\in\mathcal{X}|\alpha (\mat{x},\mat{y})=1\}$ and $\mathcal{I}(\mat{x})=\{\mat{y}|\|\mat{y}\|_2\leq \|\mat{x}\|_2\}$. A chain generated by the Metropolis-Adjusted Langevin algorithm is {\bf inwardly convergent} if 
\begin{equation}
\label{eq: inwardly convergent}
    \lim_{\|\mat{x}\|_2\rightarrow \infty} \int_{\mathcal{A}(\mat{x})\Delta \mathcal{I}(\mat{x})} t(\mat{x},\mat{y})d\mat{y}=0
\end{equation}
where $\mathcal{A}(\mat{x})\Delta \mathcal{I}(\mat{x})=\left(\mathcal{A}(\mat{x})\cup \mathcal{I}(\mat{x})\right)\setminus \left(\mathcal{A}(\mat{x})\cap \mathcal{I}(\mat{x})\right)$ is the symmetric difference.
\end{definition}
In the context of a Gaussian distribution, inward convergence states that when the norm of the current state is large accepted MCMC proposals tend towards the mean. This condition is satisfied in practice by balancing the MALA step size with the norm of the score function. For a thorough discussion see \cite{roberts1996exponential}. 

\subsection{Results}
\label{sec: convergence main results}
To achieve convergence we require that the model order $|\dict_i|$ is lower bounded by a monotone increasing function $f(i)= \Omega\left(\sqrt{i\log(i)}\right)$.  This lower bound on model order growth controls the convergence rate and influences the sequence of maximum possible thinning budgets $\{\epsilon_i\}_{i=1}^{\infty}$. 

\begin{theorem}[Decaying Thinning Budget]\label{thm: decaying budget}
Assume that the kernel $k_0$ satisfies $\mathbb{E}_{\mat{y}\sim P}\left[e^{\gamma k_0(\mat{y},\mat{y})}\right]=b < \infty$, dictionary sizes are lower bounded as $|\dict_i|\geq Cf(i)$ where $f(i)= \Omega\left(\sqrt{i\log(i)}\right)$ and specify the compression budget $\epsilon_i=\log(i)/f(i)^2$. In either of the following two cases
\begin{itemize}
\vspace{-0.03in}
\setlength\itemsep{0.3em}
    \item {\bf Case 1:} Candidate samples $\{\mat{y}_{i,l}\}^{m_i}_{l=1}$ are drawn i.i.d from the target density $p$.
    \item {\bf Case 2:} Candidate samples $\{\mat{y}_{i,l}\}^{m_i}_{l=1}$ are generated by MALA and MALA is inwardly convergent for the distantly dissipative target density $p$.
\end{itemize}
\vspace{-0.03in}
iterate $\dict_{n}={\normalfont \texttt{KSDT}}(\dict_{n-1}\cup \{\mat{x}_n\},\epsilon_n)$ of Alg. \ref{alg: outer loop} satisfies
\begin{equation}
\label{eq: decaying budget conclusion}
\mathbb{E}\left[\ksd(q_{\dict_n})^2\right] \leq C\frac{n\log(n)}{f(n)^2}.
\end{equation}
\end{theorem}

This result illustrates how both the thinning budget $\epsilon_i$ and the convergence rate depend on the model order growth $f$. A faster convergence rate requires larger $f$, and therefore both faster model order growth and a smaller thinning budget. Our result introduces a trade off between consistency (quality of representation) and complexity (number of particles retained). If we require that the dictionary size grows linearly with $f(i)\propto i$ then we recover the same $\mathcal{O}(\log(n)/n)$ asymptotic convergence rate of Theorem 1 of \cite{chen2019stein} but gain the ability to thin past samples in the dictionary. If we choose $f$ that grows at a slower rate then the sequence ${\epsilon_i}$ can decay at a slower rate so we can employ more aggressive thinning and achieve a lower-complexity representation at the cost of slower convergence. In practice we found that $\epsilon=0$ is a simple choice of thinning budget with good empirical performance. When $\epsilon=0$ thinning does not increase the KSD. Therefore the required decrease in thinning budget $\epsilon$ as $f$ grows is not a concern.

An important practical case is when we desire a fixed error representation of the density $p$. We set a fixed KSD convergence radius and then determine a constant thinning budget and the required number of steps of Algorithm \ref{alg: outer loop} to achieve a representation of $p$ with KSD error $\Delta$. 

\begin{corollary}[Constant Thinning Budget]
\label{thm: constant budget convergence corollary}
Fix target KSD convergence radius $\Delta$, assume the same conditions as Theorem \ref{thm: decaying budget}, and assume that $f(i)\propto \sqrt{i^{1+\alpha}\log(i)}$. With constant thinning budget $\epsilon=\mathcal{O}\left(\Delta^{1+\frac{1}{\alpha}}\right)$, after $n=\mathcal{O}\left(\Delta^{-\frac{1}{\alpha}}\right)$ steps the iterate $\dict_n$ produced by Algorithm \ref{alg: outer loop} satisfies
\begin{equation}
    \label{eq: contant budget corollary conclusion}
    \mathbb{E}\left[\ksd\left(q_{\dict_n}\right)^2\right]\leq C\Delta
\end{equation}
for generic constant $C$.
Equivalently, if we fix a constant thinning budget $\epsilon$, after $n=\mathcal{O}\left(\epsilon^{-\frac{2}{\alpha+1}}\right)$ steps we have
\begin{equation}
    \label{eq: contant budget corollary conclusion alternate}
 \mathbb{E}\left[\ksd\left(q_{\dict_n}\right)^2\right]\leq C\epsilon^{\frac{2}{1+\frac{1}{\alpha}}}
\end{equation}
\end{corollary}
To our knowledge, this is the first result that specifies the number of MCMC steps required to reach a fixed-error KSD neighborhood of the target density. As in Theorem \ref{thm: decaying budget} our result demonstrates a trade-off between the model order growth and convergence rate. In Theorem \ref{thm: decaying budget} we did not impose a parametric form on $f$. Here we fix the specific parametric form $f(i)\propto \sqrt{i^{1+\alpha}\log(i)}$ for clarity of exposition. Both Theorem \ref{thm: decaying budget} and Corollary \ref{thm: constant budget convergence corollary} hold in the more general case of a $V$-uniformly ergodic MCMC sampler, not just MALA. We prove this in Theorem \ref{thm: v-u ergodic convergence decaying budget} of Appendix \ref{sec: convergence appendix}.

%% file: sections/experiments.tex
\section{Experiments}\label{sec: experiments body}

In this section we present the results achieved by applying our thinning mechanism from Algorithm \ref{alg: inner loop} to several samplers, with and without the SPMCMC update rule from \eqref{eq: SPMCMC update rule}. Unlike the RBF kernel, the IMQ kernel base kernel in \ref{eq: ksd} ensures the KSD controls convergence in measure \citep{gorham2017measuring} so we use it for all experiments in the main body, then repeat all experiments with the RBF kernel in Appendix \ref{sec: experiments appendix}, drawing similar conclusions in all cases. 

In all experiments, the goal is to examine both the fidelity of the representation to $\mathbb{P}$ and the consistency/complexity tradeoff. The KSD is not sufficient for this goal because it does not consider the number of samples retained. If two dictionaries with sizes $10$ and $1000$ achieve a similar KSD, our metric should favor the dictionary of size 10 to reduce associated computational costs and storage requirements. To address this problem we define a new metric. 

From our results (Theorem \ref{thm: decaying budget}) and previous works \cite{chen2019stein} we expect the best-case KSD decay rate of $1/\sqrt{n}$ where $n$ is the number of samples. To characterize the tradeoff incurred by more complex representations we define the {\bf Normalized KSD}
\begin{equation}
    \label{eq: normalized ksd}
    \text{Normalized}\ksd(q_\dict)= \ksd(q_\dict)*\sqrt{|\dict|}
\end{equation}
which penalizes more complex representations that achieve the same KSD. If a sampler achieves the best-case KSD convergence rate of $1/\sqrt{n}$ we expect the normalized KSD to remain constant. Therefore the Normalized KSD avoids favoring larger dictionaries that are generated by longer (unthinned) sampling runs in settings where KSD evolution over time is the metric of interest. 

We consider three variants of each baseline sampler: 
\begin{itemize}[leftmargin=*]
\setlength\itemsep{0.2em}
    \item {\bf Baseline}: the baseline sampler with no thinning.
    \item {\bf KSDT-LINEAR}: Our proposed thinning method with linear dictionary growth $f(i)=i/2$.
    \item {\bf KSDT-SQRT}: Our proposed thinning method with sub-linear dictionary growth $f(i)=\sqrt{i\log(i)}$.
\end{itemize}
For simplicity we use the constant thinning budget $\epsilon=0$ in all experiments. Tuning the dictionary growth rate and thinning budget may improve results. The only hyper-parameter optimization we performed was to tune the baseline samplers for the Bayesian neural network problems.

\subsection{Goodwin Oscillator}
\begin{figure}[t]
    \centering
    \includegraphics[width=3.3in]{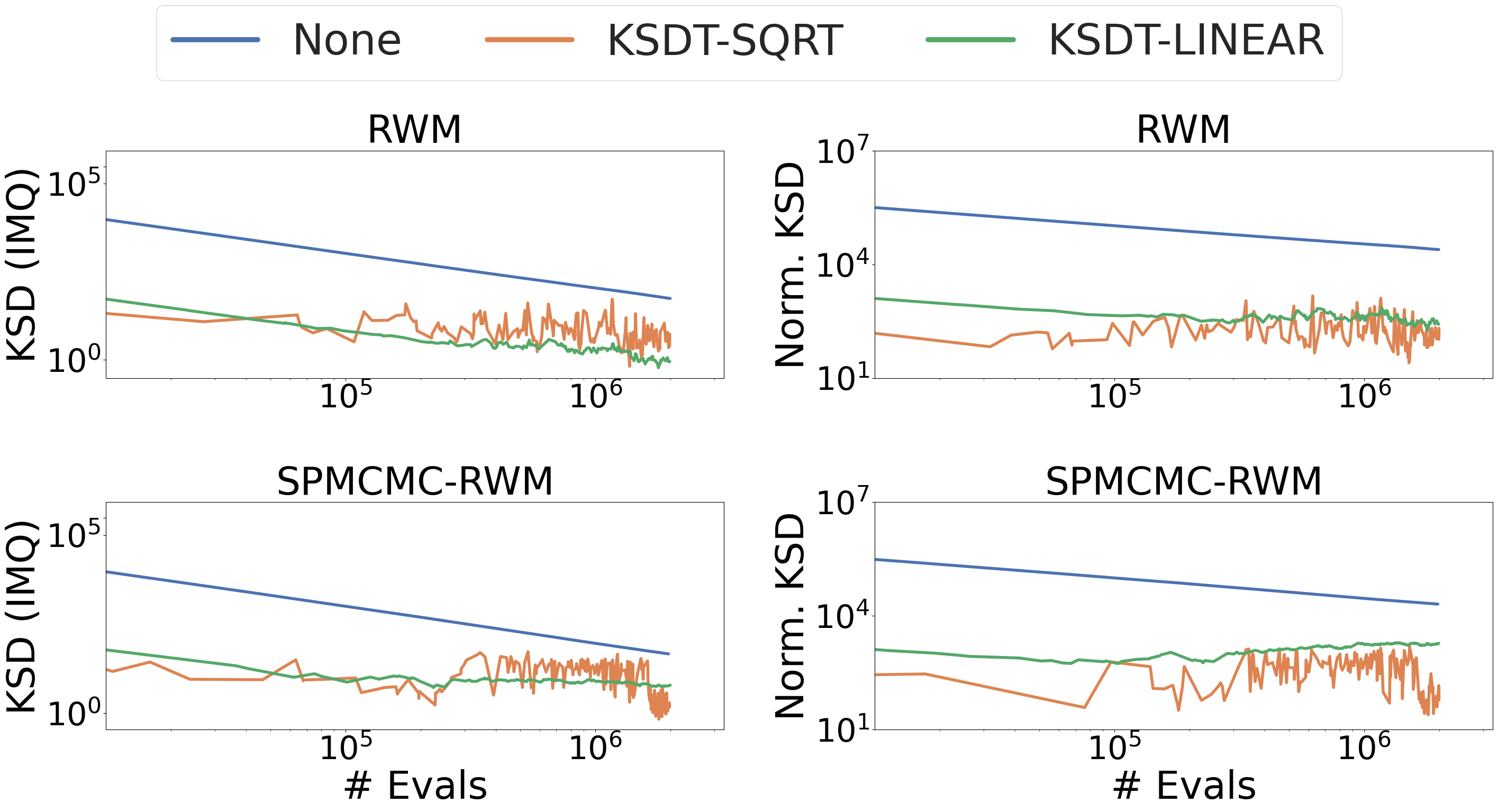}
    \caption{Goodwin oscillator problem with RWM and SPMCMC-RWM base samplers. Lower KSD and lower Normalized KSD are better. Axes are log-scale. Our KSDT-SQRT/LINEAR methods outperform the baseline methods on KSD and Normalized KSD.} 
    \label{fig: goodwin imq}
\end{figure}

The Goodwin Oscillator \citep{goodwin1965oscillatory} is a well-studied test problem for Bayesian inference of ODE parameters \citep{calderhead2009estimating,chen2019stein,riabiz2020optimal}. The task is to infer a 4-dimensional parameter that governs a genetic regulatory process specified by a system of coupled ODEs. See Appendix S5.2 of \cite{riabiz2020optimal} for details. We use Random Walk Metropolis (RWM) and Metropolis-Adjusted Langevin Algorithm (MALA) chains of length $2\times 10^6$ taken from a public data repository\footnote{https://dataverse.harvard.edu/dataset.xhtml?\\persistentId=doi:10.7910/DVN/MDKNWM} which contains the sample chains used by \cite{riabiz2020optimal}. Since chains are pre-specified and all KSD-aware sample selection methods are deterministic, we do not present error bars for this experiment. In Figure \ref{fig: goodwin imq} we report the results when applying KSD thinning to both the RWM chain and the RWM chain with the SPMCMC update rule applied. We follow the SPMCMC authors' candidate set size of $m=10$ (see \eqref{eq: SPMCMC update rule}) for this problem \citep{chen2019stein}. Our thinning methods outperform the baseline samplers on both KSD and Normalized KSD metrics. Our method with linear budget (KSDT-LINEAR) outperforms the square-root budget (KSDT-SQRT) on KSD but not normalized KSD due to the slower dictionary growth rate (smaller number of retained samples) of KSDT-SQRT. Results for the MALA and SPMCMC-MALA methods are reported in Appendix \ref{sec: goodwin appendix}. 

\subsection{Calcium Signalling Model}
\begin{figure}[t]
    \centering
    \includegraphics[width=3.3in]{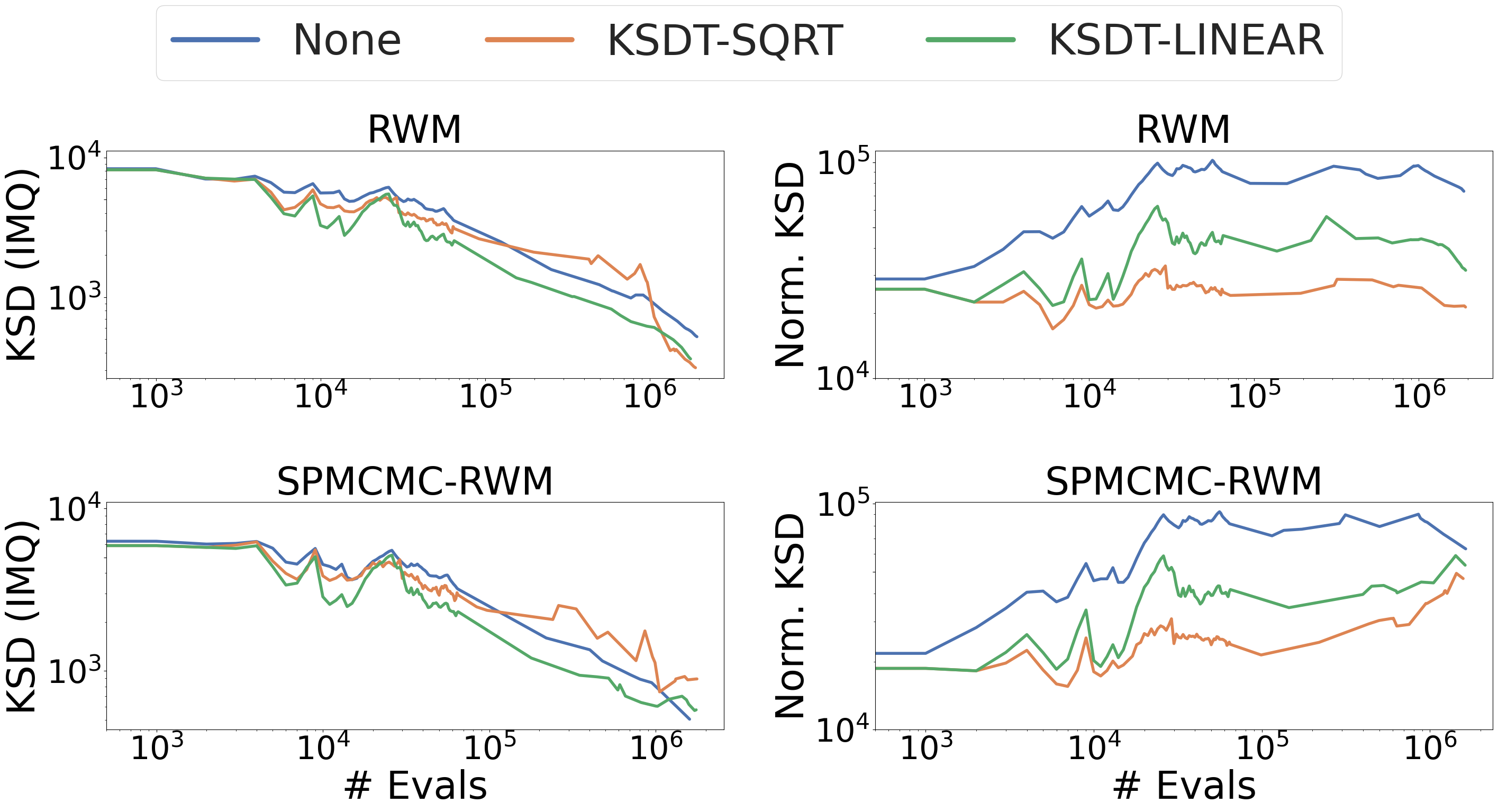}
    \caption{Calcium signalling model problem with tempered RWM and SPMCMC-RWM base samplers. Lower KSD and lower Normalized KSD are better. Axes are log-scale. Our methods outperform the baseline methods on Normalized KSD, and KSDT-LINEAR outperforms on KSD through most of the run.} 
    \label{fig: cardiac imq}
\end{figure}

We consider a calcium signaling model detailed in Appendix S5.4 of \citep{riabiz2020optimal}. The task is to infer a 38-dimensional parameter governing the signalling model. Uncertainty in the calcium signalling cascade parameter is used to propagate uncertainty to tissue-level simulations. The dataset consists of $4\times 10^6$ samples of the RWM sampler applied to a tempered posterior distribution. The samples are obtained from the same public repository as the Goodwin model samples\footnotemark[1]. We set the SPMCMC parameter $m=100$ for this problem since many samples are duplicated due to MCMC rejection. This results in approximately $10$ unique samples per step. We report KSD and Normalized KSD results in Figure \ref{fig: cardiac imq}. We observe that KSDT-LINEAR outperforms the baseline sampler on both KSD and normalized KSD in all settings except for the tail end of the SPMCMC-RWM chain. Both KSDT-Linear and KSDT-SQRT outperform the baseline sampler after normalization based on sample complexity is taken into account. 

\subsection{Bayesian Neural Network Subspace Inference}

\begin{figure}[t]
    \centering
    \includegraphics[width=3.3in]{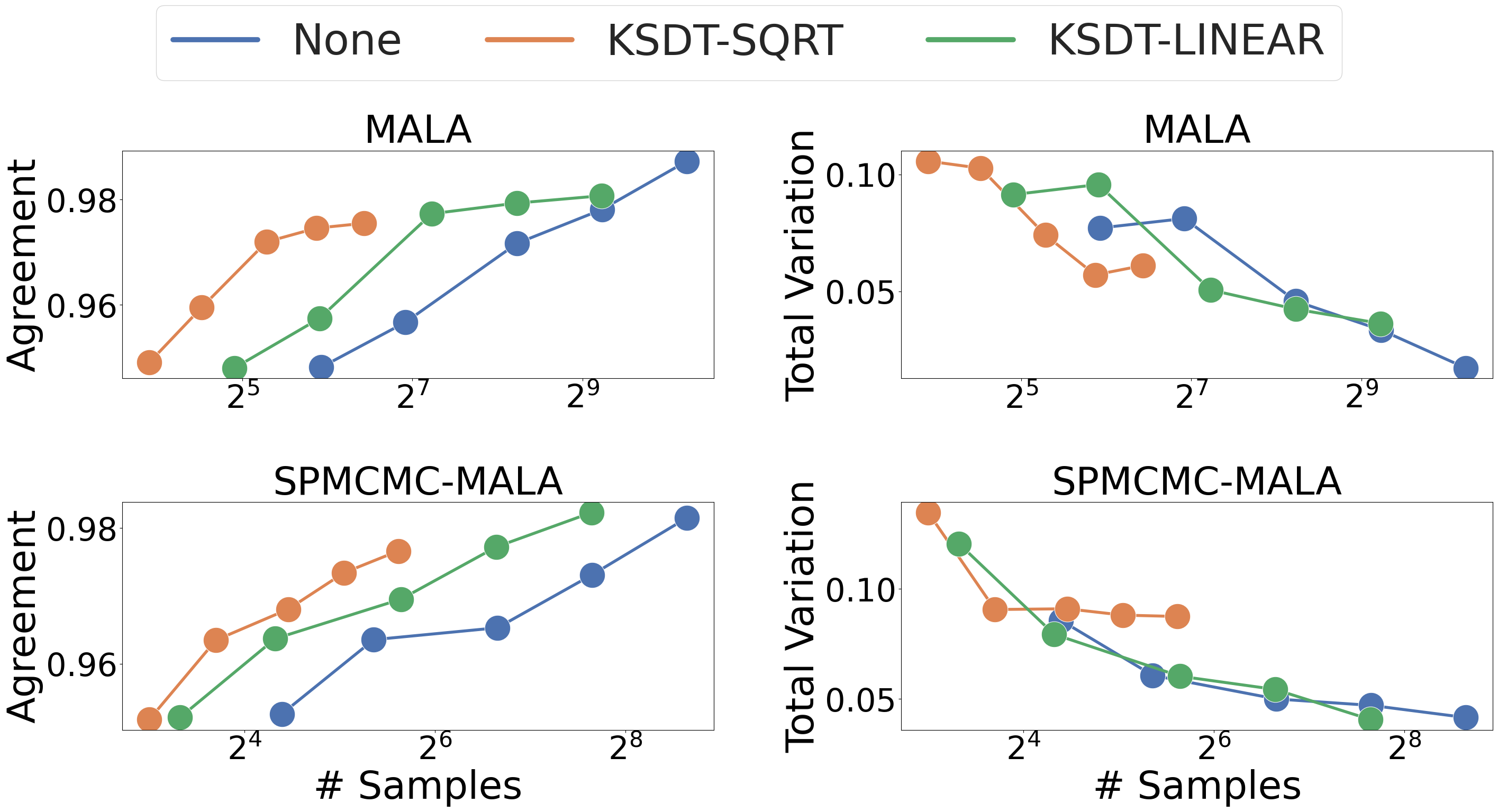}
    \caption{Agreement (left) and Total Variation (right) with the ground truth CIFAR-10 posterior predictive distribution. Higher Agreement and lower Total Variation indicate more accurate representations of the posterior predictive distribution. The x-axis is log-scale. Our KSDT-SQRT/LINEAR methods Pareto-dominate the baseline sampler on the Agreement metric and improve the low-sample Pareto frontier for the Total Variation metric.} 
    \label{fig: cifar imq agreement tv}
\end{figure}
Model complexity is a major challenge in sampling-based Bayesian deep learning. Each prediction on unseen test data requires one forward propagation per-sample so storage and inference costs grow linearly with the number of samples retained. In this section we demonstrate how our method improves the consistency vs. complexity Pareto frontier. 

MCMC mixing is slow in high dimensions, so a popular technique is to find a low-dimensional subspace and perform Bayesian sampling in that subspace \citep{constantine2016accelerating,cui2016dimension,izmailov2020subspace}. Full subspace construction details are given in Appendix \ref{sec: bnn subspace appendix}. We measure the quality of each sampling method's approximation to the predictive distribution corresponding to the true posterior. We generate the ground truth predictive distribution by running 20,000 samples of SGLD \citep{welling2011bayesian}, and follow \cite{izmailov2021bayesian} by measuring the top-1 agreement and total variation with respect to the ground truth predictive distribution. 
Total variation is
\begin{equation}
    \label{eq: total variation}
    \frac{1}{n}\sum_{i=1}^n \frac{1}{2}\sum_{j=1}^c\left\lvert{p_1(\mat{y}_i=j|\mat{x}_i)-p_2(\mat{y}_i=j|\mat{x}_i)}\right\rvert
\end{equation}
where $c$ is the number of classes. Higher agreement and lower total variation indicate higher-fidelity approximations to the ground truth predictive distribution. All results in this section are the mean over 5 chains. More details and additional experiments are given in Appendix \ref{sec: experiments appendix}.

\begin{figure}[t]
    \centering
    \includegraphics[width=3.3in]{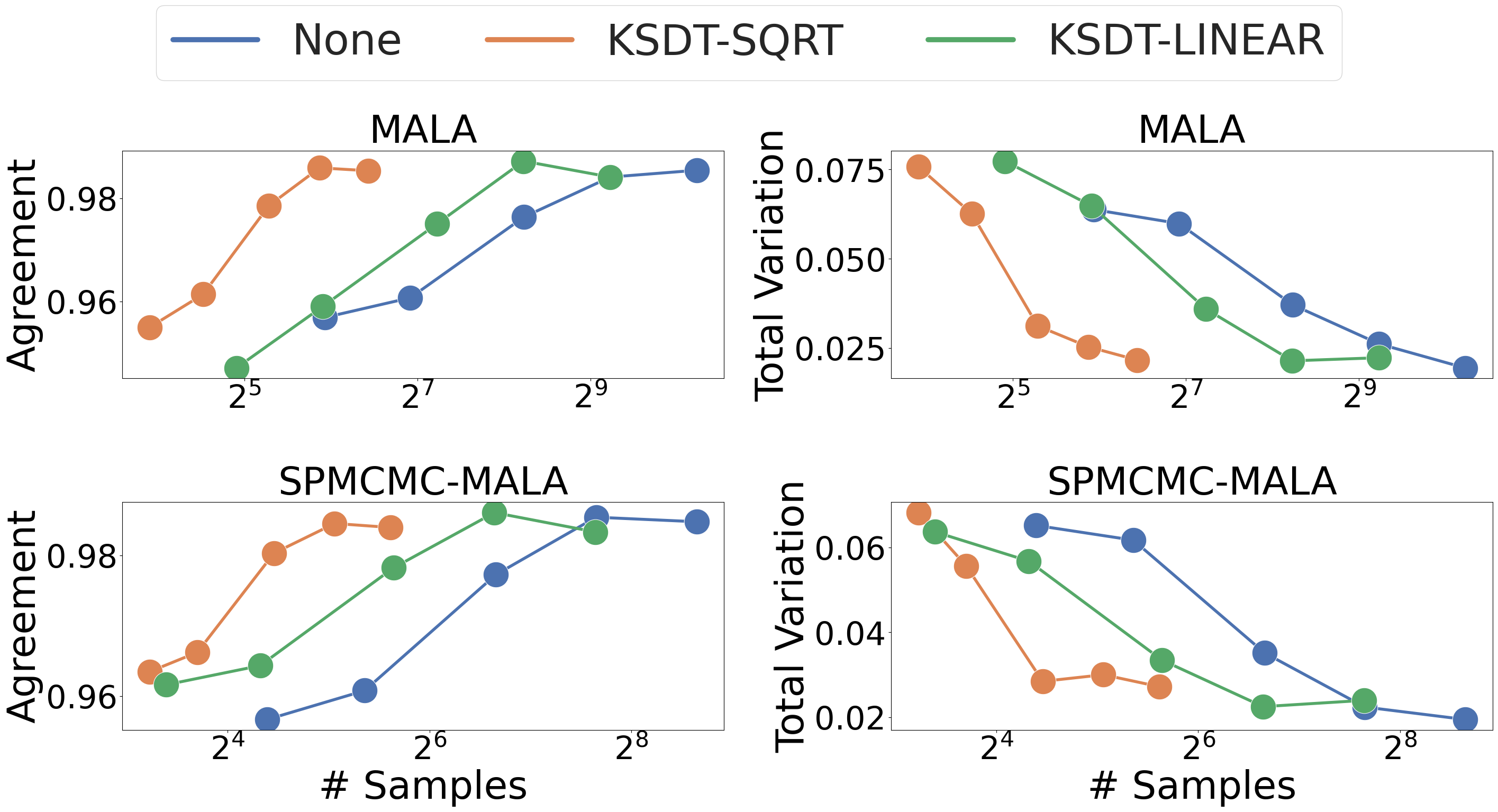}
   \caption{Agreement (left) and Total Variation (right) with the ground truth IMDB posterior predictive distribution. Higher Agreement and lower Total Variation indicate more accurate representations of the posterior predictive distribution. The x-axis is log-scale. Our KSDT-SQRT/LINEAR methods Pareto-dominate the baseline sampler on both the Agreement and Total Variation metrics.}
    \label{fig: imdb imq agreement tv}
\end{figure}

\paragraph{CIFAR-10 Classification}
The task is 10-class image classification, and we use a ResNet-20 with Filter Response Normalization from \citep{izmailov2021bayesian} as our base model. Our results in Figure \ref{fig: cifar imq agreement tv} demonstrate that our online thinning method outperforms both the baseline sampler and SPMCMC-based samplers on the agreement metric. Improvement is mixed on the total variation metric, where our methods outperform existing methods in the low-sample regime, but have similar performance to existing methods once the model complexity is high.

\paragraph{IMDB Sentiment Prediction}
The task is two-class sentiment prediction and we use a CNN-LSTM from \citep{izmailov2021bayesian} as our base model. Our results in Figure \ref{fig: imdb imq agreement tv} demonstrate that all of our online thinning methods, in particular KSDT-SQRT, improve the Pareto frontier of the corresponding baseline sampler on both metrics. This conclusion holds across all sample regimes.

\subsection{Automatic Adaptation to Target Complexity}\label{sec: adaptivity}
\begin{figure}
    \centering
    \includegraphics[width=\linewidth]{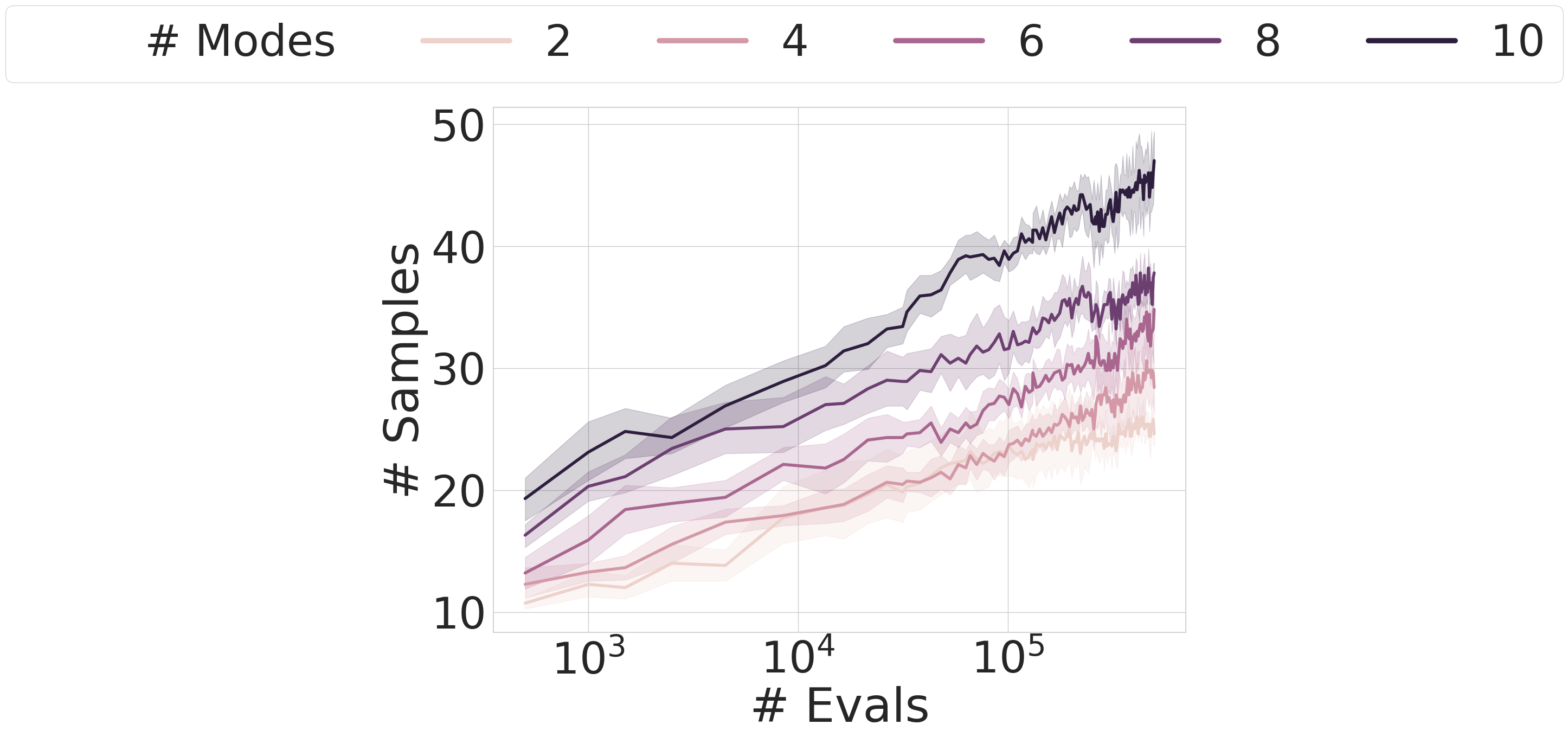}
    \caption{Thinning without mandatory dictionary size growth $f(i)=10$. Our algorithm automatically adapts to the number of modes by retaining 40 samples for a 10-mode mixture and 24 for a 4-mode mixture.}
    \label{fig: complexity adaptation}
\end{figure}
We have studied our KSD-Thinning algorithm with mandatory dictionary growth to ensure asymptotic consistency via Theorem \ref{thm: decaying budget}. In this section we observe an interesting property of our algorithm: {\it If we do not enforce model order growth, our thinning algorithm automatically adapts to the target distribution complexity}. We target equally weighted Gaussian mixture models with an increasing number of modes and set $f(t)=10$. We match the setting of Theorem \ref{thm: decaying budget} and draw samples directly from the true distribution, applying the SPMCMC update rule with $m=5$. Figure \ref{fig: complexity adaptation} shows that the number of retained samples increases with the number of modes.

%% file: sections/conclusion.tex
\section{Conclusion}

In this paper we presented an online thinning method that produces a compressed representation of a target distribution. Our work enables MCMC algorithms to directly target complexity-aware representations during, and can be applied to MCMC sampling scheme when gradients are available. We demonstrated the broad applicability of our method by comparing it to many baseline samplers in several problem settings, and found that it often improves model fit and reduces model complexity.



%% file: sections/convergence_appendix.tex
\section{Convergence Results}\label{sec: convergence appendix}
We provide guarantees of convergence when the sample chain $\seq$ is generated by SPCMCMC-style samplers. Our convergence results are provided in expectation, where the expectation  $\mathbb{E}$ is taken with respect to the distribution underlying the sample generating process for $\seq$. Specifically, the sample stream $\seq$ is a sequence of random variables from a time-invariant distribution. Each realization of $\seq$ is either a single MCMC sample chain, or the concatenation of several MCMC sample chains according to the SPMCMC update rule given in \eqref{eq: SPMCMC update rule}.

The two different sampling mechanisms that we consider are (1) i.i.d. sampling from the target distribution and (2) V-uniformly ergodic MCMC sampling. While we do not expect to directly draw samples from the true distribution, the i.i.d. sampling analysis is similar to the MCMC analysis and provides a simpler starting point. The decaying budget result from Theorem \ref{thm: decaying budget} is proven in two parts. Theorem \ref{thm: iid convergence decaying budget} gives the desired result for i.i.d. sampling and Corollary \ref{thm: mala convergence decaying budget} gives the desired result for the MALA sampler. A more general result for V-uniformly ergodic MCMC samplers is given in Theorem \ref{thm: v-u ergodic convergence decaying budget}. The constant-budget result is given for i.i.d. sampling in Corollary \ref{thm: iid constant budget convergence corollary} and for the MALA sampler in Corollary \ref{thm: mala constant budget convergence corollary}.

\begin{table}
\begin{center}
\begin{tabular}{ c| l }
 Symbol & Definition  \\ \hline
 $k_0$ & Stein kernel as defined in \eqref{eq: stein kernel}  \\  \hline
 $\mathcal{K}_0$ & RKHS induced by Stein kernel \\  \hline
 $\ksd$ & KSD as defined in \eqref{eq: ksd}  \\  \hline
 $\mathcal{Y}_i$ & search space for best next point to add\\  \hline
 $\mat{x}_n$ & new point added to dictionary $\dict_{t-1}$ before projection step\\ \hline
 $S_n$ & ``size" of search space $Y_i$  \\  \hline
 $h_n$ & Optimal RKHS update given search space $\mathcal{Y}_i$\\  \hline
 $\delta$ & suboptimality incurred during search for KSD-optimal point\\ \hline
 $r_n$ & arbitrary positive constant\\ \hline
 $m_n$ & size of new point candidate set\\ \hline
 $\{\mat{y}_{n,i}\}$ & candidate set of MCMC samples\\ \hline
 $\gamma,b$ & constants used to bound exponentiated kernel\\ \hline

\end{tabular}
\end{center}
\caption{Symbols used for convergence proofs}
\end{table}




The following technical lemma imposes constraints on the sample stream $\seq$. We will partition $\seq$ into a sequence of batches $\mathcal{Y}_i$ and select the KSD-optimal element from each batch. The lemma below is stated for generic constants $S_i,r_i$ and search sets $\mathcal{Y}_i$ to accommodate two cases (1) $\seq$ is a stream of i.i.d samples from the target distribution (2) $\seq$ is a stream of MCMC samples.

\begin{lemma}[General Recursion Lemma, modified from Theorem 5 in \cite{supchen2019stein}]
\label{lem: general thinned recursion lemma}
Let $\mathcal{X}$ be the domain of $p$. Fix $n\in \mathbb{N}$. Assume that for all $j\leq n$ the sequence of points $\{\mat{x}_i\}_{i=1}^j$, where $\mat{x}_i$ the greedy KSD-optimal point selected from the search space $\mathcal{Y}_i\subset \mathcal{X}$, satisfies
\begin{equation}
\label{eq: general thinned recursion lemma premise}
\frac{k_0(\mat{x}_j,\mat{x}_j)}{2}+\sum_{\mat{x}_i\in\dict_{j-1}}k_0(\mat{x}_i,\mat{x}_j)\leq \frac{\delta}{2}+\frac{S_j^2}{2}+\inf_{\mat{x}\in \mathcal{Y}_j}\sum_{\mat{x}_i\in\dict_{j-1}}k_0(\mat{x}_i,\mat{x}).
\end{equation}
Then for any constants $\delta>0$, $\{S_i\geq 0\}_{i=1}^n$, $\{r_i>0\}_{i=1}^n$ the thinned dictionary $\dict_{n}=KSDT(\dict_{n-1}\cup \{\mat{x}_n\},\epsilon_n)$ satisfies
\begin{equation}
    \label{eq: general thinned recursion lemma conclusion}
\ksd(q_{\dict_n})^2 \leq \sum_{i=1}^{n-1} \left( \frac{\delta+S_i^2+r_i\|h_i\|_{\mathcal{K}_0}^2}{\left(|\dict_{i-1}|+1\right)^2}+ \epsilon_i \right)\left(\prod_{j=i}^{n-1} \frac{|\dict_j|}{|\dict_j|+1}\right)^2 \left( \prod_{j=i}^{n-1} \left(1+\frac{1}{r_{j+1}}\right)\right)  
\end{equation}
where $h_i$ is any element of the Stein RKHS corresponding to points in the convex hull of $\mathcal{Y}_i$.
\end{lemma}
The main distinctions between Lemma \ref{lem: general thinned recursion lemma} and Theorem 5 of \cite{supchen2019stein} is that we consider additional thinning through Algorithm \ref{alg: inner loop}. This thinning operation introduces the budget parameter $\epsilon_i$ (potential KSD increase) and the dictionary size $|\dict_j|$ since the dictionary size is not a linear monotonically increasing function of the time step as in \citep{supchen2019stein}

\paragraph{Proof}

The condition in \eqref{eq: general thinned recursion lemma premise} assumes that the point sequence $\{\mat{x}_j\}$ has bounded suboptimality $\delta$ with respect to KSD minimization at each step. Then the conclusion in \eqref{eq: general thinned recursion lemma conclusion} gives an upper bound on the squared KSD at step $n$. The general constants $\{S_i\},\{r_i\}$ will be instantiated based on the choice of search spaces $\{\mathcal{Y}_i\}$ and the choice of point generation method (MCMC, deterministic optimization, etc.).

First we will present a bound on the 1-step unthinned iterate. Next we account for KSD loss due to thinning to achieve a 1-step bound for our algorithm. Finally we recursively apply this bound to achieve the recurrence relation in \eqref{eq: general thinned recursion lemma conclusion}. 

First we provide a 1-step bound for the unthinned iterate $\tilde\dict_n$ by adapting the proof style of Theorem 5 in Appendix A.1 of \cite{supchen2019stein} to suit our notation. 

\begin{align}
        |\tilde\dict_n|^2\ksd(q_{\tilde\dict_n})^2 &= \sum_{\mat{x}_i\in\tilde\dict_n}\sum_{x_j\in \tilde\dict_n}  k_0(\mat{x}_i,\mat{x}_j)\nonumber\\
        &=|\dict_{n-1}|^2\ksd(q_{\dict_{n-1}})^2+k_0(\mat{x}_n,\mat{x}_n)+2\sum_{\mat{x}_i\in\dict_{n-1}}k_0(\mat{x}_i,\mat{x}_n)\nonumber\\
        &\leq|\dict_{n-1}|^2\ksd(q_{\dict_{n-1}})^2+\delta+S_n^2+2\inf_{\mat{x}\in\mathcal{Y}_j}\sum_{\mat{x}_i\in\dict_{n-1}}k_0(\mat{x}_i,\mat{x}) \label{eq: intermediate ksd lastline}
\end{align}
The equality follows from the definition of the KSD in \eqref{eq: ksd}. The second equality follows from the definition of the KSD again as we separate the elements in the final row and final column of the kernel matrix to get the second two terms. The last inequality is the direct application of the premise of Lemma~\ref{lem: general thinned recursion lemma}, which ensures bounded suboptimality of the point $\mat{x}_n$, to the last two terms on the right-hand side of the middle equality in \eqref{eq: intermediate ksd lastline}. Multiply the premise by 2 for direct application.

We apply the following result from Equation (11) of \cite{supchen2019stein}:
\[
2\inf_{x\in\mathcal{Y}_j}\sum_{\mat{x}_i\in\dict_{n-1}}k_0(\mat{x}_i,\mat{x})\leq r_n\|h_n\|^2_{\mathcal{K}_0}+\frac{\ksd(q_{\dict_{n-1}})}{r_n}
\]
to the last term on the right-hand side of \eqref{eq: intermediate ksd lastline} to get
\begin{equation}
        |\tilde\dict_n|^2\ksd(q_{\tilde\dict_n})^2 \leq|\dict_{n-1}|^2\left(1+\frac{1}{r_n}\right)\ksd(q_{\dict_{n-1}})^2+\delta+S_n^2+r_n\|h_n\|_{\mathcal{K}_0}^2.
\end{equation}
To recover the un-thinned KSD we divide both sides by both sides by $|\tilde\dict_n|^2=\left(|\dict_{n-1}|+1\right)^2$ to get 
\begin{equation}
        \ksd(q_{\tilde\dict_n})^2 \leq \frac{|\dict_{n-1}|^2}{\left(|\dict_{n-1}|+1\right)^2}\left(1+\frac{1}{r_n}\right)\ksd(q_{\dict_{n-1}})^2+\frac{\delta+S_n^2+r_n\|h_n\|_{\mathcal{K}_0}^2}{\left(|\dict_{n-1}|+1\right)^2}\\
\end{equation}
We replace the un-thinned KSD $\ksd(q_{\tilde\dict_n})^2$ with the thinned KSD $\ksd(q_{\dict_n})^2$ and apply the stopping criterion of Algorithm ~\ref{alg: inner loop} to establish the recurrence relationship
\begin{equation}
\ksd(q_{\dict_n})^2 \leq \frac{|\dict_{n-1}|^2}{\left(|\dict_{n-1}|+1\right)^2}\left(1+\frac{1}{r_n}\right)\ksd(q_{\dict_{n-1}})^2+\frac{\delta+S_n^2+r_n\|h_n\|_{\mathcal{K}_0}^2}{\left(|\dict_{n-1}|+1\right)^2} +\epsilon_n.
\end{equation}

This recurrence may be unrolled backwards in time to write
\begin{equation}
\ksd(q_{\dict_n})^2 \leq \sum_{i=1}^{n} \left( \frac{\delta+S_i^2+r_i\|h_i\|_{\mathcal{K}_0}^2}{\left(|\dict_{i-1}|+1\right)^2}+ \epsilon_i \right)\left(\prod_{j=i}^{n-1} \frac{|\dict_j|}{|\dict_j|+1}\right)^2 \left( \prod_{j=i}^{n-1} \left(1+\frac{1}{r_{j+1}}\right)\right)  
\end{equation}
which upper bounds the error incurred by each step of the non-parametric approximation to the target density $p$, as stated in Lemma \ref{lem: general thinned recursion lemma}. \hfill $\blacksquare$

Next we present Lemma \ref{lem: thinned spmcmc recursion lemma} which is a consequence of Lemma \ref{lem: general thinned recursion lemma} when the KSD-optimal point selection is no longer generic, but instead the specific outcome of the SPMCMC update rule from \eqref{eq: SPMCMC update rule} applied to the candidate discrete search space. For the following lemma, we assume that we are given a set of candidate samples from the target density $p$. We partition those samples into candidate sets $\mathcal{Y}_i=\{\mat{y}_{i,l}\}_{l=1}^{m_i}$ of size $m_i$. Then the sample stream $\seq=\{\mat{x}_i\}_{i=1}^\infty$ is generated by the SPMCMC update from \eqref{eq: SPMCMC update rule}. The following lemma is a modified form of Theorem 6 from \citep{supchen2019stein}.

\begin{lemma}[Pruned Recursion with SPMCMC Update Rule]
\label{lem: thinned spmcmc recursion lemma}
Assume the same setup as Lemma ~\ref{lem: general thinned recursion lemma}. Using the SPMCMC update rule from \eqref{eq: SPMCMC update rule} the iterate $\dict_{n}=KSDT(\dict_{n-1}\cup \{\mat{x}_n\},\epsilon_n)$ satisfies
\begin{equation}
\label{eq: iid recursion conclusion}
\ksd(q_{\dict_n})^2 \leq \sum_{i=1}^{n} \left( \frac{S_i^2+r_i\|h_i\|_{\mathcal{K}_0}^2}{\left(|\dict_{i-1}|+1\right)^2}+ \epsilon_i \right)\left(\prod_{j=i}^{n-1} \frac{|\dict_j|}{|\dict_j|+1}\right)^2 \left( \prod_{j=i}^{n-1} \left(1+\frac{1}{r_{j+1}}\right)\right)  
\end{equation}
\end{lemma}
\paragraph{Proof}
In this proof we will apply Lemma ~\ref{lem: general thinned recursion lemma} to this specific case by instantiating a specific constant $\delta$ related to the search procedure and bounding the search space $\mathcal{Y}_i$ with respect to $S_i$.

First we truncate and redefine the search set $\mathcal{Y}_n$ by restricting our attention to regions with kernel values bounded by $S_n^2$: $\mathcal{Y}_n = \{\mat{x}\in\mathcal{Y}_n\:|\:k_0(\mat{x},\mat{x})\leq S_n^2\}$. Then we note that the update rule in Equation~\ref{eq: SPMCMC update rule} satisfies the premise (\eqref{eq: general thinned recursion lemma premise}) in Lemma~\ref{lem: general thinned recursion lemma} with $\delta=0$ because the infimum is a search over a finite set of points, and therefore the KSD suboptimality $\delta$ of the search procedure is $\delta=0$.
    \begin{align}
\frac{k_0(\mat{x}_n,\mat{x}_n)}{2}+\sum_{\mat{x}_i\in\dict_{n-1}}k_0(\mat{x}_i,\mat{x}_n)&=\inf_{\mat{x_n}\in\mathcal{Y}_n}\frac{k_0(\mat{x}_n,\mat{x}_n)}{2}+\sum_{\mat{x}_i\in\dict_{n-1}}k_0(\mat{x}_i,\mat{x}_n) \nonumber\\
&\leq\frac{S_n^2}{2}+\inf_{\mat{x}\in \mathcal{Y}_n}\sum_{\mat{x}_i\in\dict_{n-1}}k_0(\mat{x}_i,\mat{x}) \label{eq: plug in discrete infimum}
\end{align}
The first equality follows from the optimality condition of the update rule (select the best point). The second line follows from the truncation criterion of $\mathcal{Y}_n$. Finally we can apply Lemma \ref{lem: general thinned recursion lemma} with $\delta=0$ to achieve the desired conclusion in ~\eqref{eq: iid recursion conclusion}. This concludes the proof of Lemma \ref{lem: thinned spmcmc recursion lemma}. \hfill $\blacksquare$

Lemma \ref{lem: general thinned recursion lemma} is a stepping stone to Lemma \ref{lem: thinned spmcmc recursion lemma}. We use Lemma \ref{lem: thinned spmcmc recursion lemma} to establish convergence of Algorithm \ref{alg: outer loop} in the i.i.d. sampling setting of Theorem \ref{thm: iid convergence decaying budget} by bounding the summation in \eqref{eq: iid recursion conclusion} and therefore ensuring KSD convergence.

To achieve convergence we will require that the model order $|\dict_i|$ is lower bounded by a monotone increasing function $f(i)\in \Omega\left(\sqrt{i\log(i)}\right)$. The lower bound on model order growth controls the convergence rate and influences maximum possible thinning budget $\{\epsilon_i\}$. This lower bound contrasts with the standard linear $|\dict_i|=i$ growth rate for MCMC or i.i.d. sampling and the linear $f(i)\in\mathcal{O}(i)$ growth rate of \cite{supchen2019stein}.

\begin{theorem}[i.i.d Sampling and Decaying Thinning Budget]\label{thm: iid convergence decaying budget}
Assume the same conditions as Lemma \ref{lem: thinned spmcmc recursion lemma}. Further assume that the kernel $k_0$ satisfies $\mathbb{E}_{\mat{y}\sim P}\left[e^{\gamma k_0(\mat{y},\mat{y})}\right]=b < \infty$, dictionary sizes are lower bounded as $|\dict_i|\geq Cf(i)$ where $f(i)= \Omega\left(\sqrt{i\log(i)}\right)$ and that the compression budget $\{\epsilon_i\}$ is $\epsilon_i=\log(i)/f(i)^2$. Finally assume that the candidate samples $\{\mat{y}_{i,l}\}^{m_i}_l$ are drawn i.i.d from the target density $p$. Then the iterate $\dict_{n}=KSDT(\dict_{n-1}\cup \{\mat{x}_n\},\epsilon_n)$ of Algorithm \ref{alg: outer loop} satisfies
\begin{equation}
\label{eq: iid convergence decaying budget conclusion}
\mathbb{E}\left[\ksd(q_{\dict_n})^2\right] \leq C\frac{n\log(n)}{f(n)^2}.
\end{equation}
\end{theorem}

\paragraph{Proof}
To complete our proof of convergence first we will split the recurrence from Lemma ~\ref{lem: general thinned recursion lemma} into two terms, one corresponding to the sampling error and one corresponding to the thinning error. Then we will bound each individually by selecting specific values of $\{r_i\},\{S_i\}$ and $\{m_i\}$ and applying a bound for $\|h_i\|_{\mathcal{K}_0}$.\\

First we apply standard log-sum-exp bound  as in \cite{supchen2019stein} the part related to the sampling error to write
\begin{equation}
    \label{eq: r product}
    \prod_{j=i}^{n-1}\left(1+\frac{1}{r_{j+1}}\right)\leq \exp\left(\sum_{j=1}^n \frac{1}{r_j}\right).
\end{equation}
We substitute \eqref{eq: r product} into the conclusion of Lemma \ref{lem: thinned spmcmc recursion lemma} to get
\begin{equation}
\label{eq: initial recursion with r}
\begin{split}
\mathbb{E}\left[\ksd(q_{\dict_n})^2\right] &\leq \mathbb{E}\left[\exp\left(\sum_{j=1}^n \frac{1}{r_j}\right)\sum_{i=1}^{n} \left( \frac{S_i^2+r_i\|h_i\|_{\mathcal{K}_0}^2}{\left(|\dict_{i-1}|+1\right)^2}+ \epsilon_i \right)\left(\prod_{j=i}^{n-1} \frac{|\dict_j|}{|\dict_j|+1}\right)^2\right]\\
\end{split}
\end{equation}
For now we ignore the leading exponential and decompose the summation into two terms related to the sampling and compression-based error, respectively, as:
\begin{equation}
\label{eq: T1 T2 introduction}
\begin{split}
\sum_{i=1}^{n}& \left( \frac{S_i^2+r_i\|h_i\|_{\mathcal{K}_0}^2}{\left(|\dict_{i-1}|+1\right)^2}+ \epsilon_i \right)\left(\prod_{j=i}^{n-1} \frac{|\dict_j|}{|\dict_j|+1}\right)^2\\
&=\underbrace{\sum_{i=1}^{n} \left( \frac{S_i^2+r_i\|h_i\|_{\mathcal{K}_0}^2}{\left(|\dict_{i-1}|+1\right)^2}\right)\left(\prod_{j=i}^{n-1} \frac{|\dict_j|}{|\dict_j|+1}\right)^2}_{T_1}+
\underbrace{\sum_{i=1}^{n} \epsilon_i\left(\prod_{j=i}^{n-1} \frac{|\dict_j|}{|\dict_j|+1}\right)^2}_{T_2}
\end{split}
\end{equation}
The term $T_1$ is the bias incurred at each step of the unthinned point selection scheme. The term $T_2$ is the bias incurred by the thinning operation carried out at each step.

\paragraph{Bound on $T_1$} The term $T_1$ is the bias incurred by the point sampling scheme.
\begin{equation}
\label{eq: bound computation}
\begin{split}
    \mathbb{E}\Bigg[\sum_{i=1}^{n} \left( \frac{S_i^2+r_i\|h_i\|_{\mathcal{K}_0}^2}{\left(|\dict_{i-1}|+1\right)^2}\right)&\left(\prod_{j=i}^{n-1} \frac{|\dict_j|}{|\dict_j|+1}\right)^2 \Bigg] \\
    &=\mathbb{E}\left[\sum_{i=1}^{n} \left( \frac{S_i^2+r_i\|h_i\|_{\mathcal{K}_0}^2}{\left(|\dict_{n-1}|+1\right)^2}\right)\left(\prod_{j=i}^{n-1} \frac{|\dict_j|}{|\dict_{j-1}|+1}\right)^2 \right] \\
    &\leq\mathbb{E}\left[\sum_{i=1}^{n} \frac{S_i^2+r_i\|h_i\|_{\mathcal{K}_0}^2}{\left(|\dict_{n-1}|+1\right)^2}\right]\\
    &C\sum_{i=1}^{n} \frac{S_i^2+r_i\mathbb{E}\left[\|h_i\|_{\mathcal{K}_0}^2\right]}{f(n)^2}
\end{split}
\end{equation}
The first equality is a re-arrangement of the denominators that pulls in the term $\frac{1}{(|\dict_{i-1}|+1)^2}$. The second inequality is a consequence of the fact that $|\dict_j|\leq |\dict_{j-1}|+1$ so each product of the dictionary order ratios is at most 1. The final term is based on the assumption of dictionary model order growth, i.e., $|\dict_{n}|\geq f(n)$. From Appendix A, Equation (17) of \cite{supchen2019stein} we can apply truncated kernel mean embeddings and the fact that the points are drawn i.i.d. from the true distribution $P$ to bound $\mathbb{E}\left[\|h_i\|_{\mathcal{K}_0}^2\right]$
by
\begin{equation}
    \label{eq: h bound iid}
    \mathbb{E}\left[\|h_i\|_{\mathcal{K}_0}^2\right]\leq\frac{4b}{\gamma}e^{-\frac{\gamma}{2}S_i^2}+\frac{4}{m_j}S_i^2
\end{equation}
which relies on the assumption that $k_0$ satisfies $\mathbb{E}_{\mat{y}\sim P}\left[e^{\gamma k_0(\mat{y},\mat{y})}\right]=b < \infty$. Therefore
\begin{align}
\label{eq: apply h bound iid decaying budget}
    \mathbb{E}\Bigg[\sum_{i=1}^{n} \left( \frac{S_i^2+r_i\|h_i\|_{\mathcal{K}_0}^2}{\left(|\dict_{i-1}|+1\right)^2}\right)&\left(\prod_{j=i}^{n-1} \frac{|\dict_j|}{|\dict_j|+1}\right)^2\Bigg]\nonumber\\
    &\leq C\frac{1}{f(n)^2}\sum_{i=1}^{n} \left((1+\frac{r_i}{m_i})S_i^2+r_ie^{-\frac{\gamma}{2}S_i^2}\right)
\end{align}
The inequality is an application of \eqref{eq: h bound iid} to the final conclusion of \eqref{eq: bound computation} followed by an absorption of the constants into $C$. Note that $C\propto \exp{\sum_{i=1}^n \frac{1}{r_i}}$. We follow \cite{supchen2019stein} and select and choose $S_i=\sqrt{\frac{2}{\gamma}\log(n\wedge m_i)}$ and $r_i=n$. The selection $r_i=n$ is necessary to ignore the leading exponential $\exp\left(\sum_{j=1}^n \frac{1}{r_j}\right)$ and render that term constant. The choices of $r_i$ and $S_i$ are artefacts of the analysis, but do not affect the practical algorithm. Set $m_i=n$ to get

\begin{align}
   \mathbb{E}\Bigg[\sum_{i=1}^{n} \left( \frac{S_i^2+r_i\|h_i\|_{\mathcal{K}_0}^2}{\left(|\dict_{i-1}|+1\right)^2}\right)&\left(\prod_{j=i}^{n-1} \frac{|\dict_j|}{|\dict_j|+1}\right)^2\Bigg]\nonumber\\
    &\leq C\frac{1}{f(n)^2}\sum_{i=1}^{n} \left(\log(n)+1\right)\nonumber\\
    &\leq C\frac{n\log(n)}{f(n)^2}. \label{eq: final T1 bound}
\end{align}
The first inequality is the conclusion of the previous equation with $S_i$ and $r_i$ evaluated and all constants absorbed. The second inequality holds because the summand does not depend on the index so we can multiply by $n$. This completes the bound on $T_1$ subject to the dictionary growth constraint.

\paragraph{Bound on $T_2$}
The term $T_2$ is the bias incurred by the thinning operation. We will tune the budget schedule $\{\epsilon_i\}$ so that $T_2$ tends to 0 at the same rate as $T_1$. The goal is to keep epsilon as large as possible in order to retain as few points as possible while preserving the rate of posterior contraction. Noticeably, we do not need this term to converge any faster than $n\log n/f(n)^2$. The only dependence on the point stream $\seq$ comes from $|\dict_j|$, and we directly bound this term by observing that $|\dict_j|\leq j$ so 
\[\frac{|\dict_j|}{|\dict_{j+1}|}\leq \frac{j}{j+1}.\] 
Therefore, returning to $T_2$ in \eqref{eq: T1 T2 introduction}, we have 
\begin{align}
\mathbb{E}\left[\sum_{i=1}^{n} \epsilon_i\left(\prod_{j=i}^{n-1} \frac{|\dict_j|}{|\dict_j|+1}\right)^2 \right] &\leq \sum_{i=1}^{n} \epsilon_i\left(\prod_{j=i}^{n-1} \frac{j}{j+1}\right)^2 \nonumber\\
&= \sum_{i=1}^{n} \epsilon_i\frac{i^2}{n^2} \nonumber\\
&= \frac{1}{n^2}\sum_{i=1}^{n} \epsilon_i i^2 
\end{align}
Therefore the goal is to choose $\epsilon_i$ satisfying (up to generic constants)
\begin{align}
\label{eq: T2 sufficient condition}
\frac{1}{n^2}\sum_{i=1}^n\epsilon_i i^2 &\leq \frac{n\log n}{f(n)^2}\nonumber\\
\iff \sum_{i=1}^n \epsilon_i i^2&\leq \frac{n^3\log n}{f(n)^2} 
\end{align}

The first line above states that $T_1$ and $T_2$ converge at the same rate. The second line simply multiplies the first inequality by $n^2$. Therefore, to satisfy the condition on the right-hand side of the previous expression, one can select $\epsilon_i$ according to the growth condition associated with the posterior contraction rate of the sampled process, i.e., the upper bound on $T_1$ in \eqref{eq: final T1 bound vue}. According to this rate, the model complexity increases at least according to the growth rate  $f(i)$. We set $\epsilon_i = \log(i)/f(i)^2$ and demonstrate that \eqref{eq: T2 sufficient condition} is met:
\begin{equation}
\label{eq: final T2 bound}
\begin{split}
\sum_{i=1}^n \epsilon_i i^2 &= \sum_{i=1}^n \frac{i^2\log(i)}{f(i)^2} \\
&\leq \sum_{i=1}^n \frac{n^2\log(n)}{f(n)^2} \\
&= \frac{n^3\log(n)}{f(n)^2}\\
\end{split}
\end{equation}
The first line is the evaluation of $\epsilon_i$. The inequality in the second line holds because $f(i)\leq i$ so the summand is increasing in $i$. After we remove the dependence on the index $i$ we multiply by the maximum summation index $n$ to achieve the desired result. This concludes the bound of $T_2$.

Finally we can substitute the bound for $T_1$ (\eqref{eq: final T1 bound}) and the bound for $T_2$ (\eqref{eq: final T2 bound}) into \eqref{eq: T1 T2 introduction} to get 

\begin{equation}
\label{eq: T1 T2 bound substitution}
\mathbb{E}\left[\sum_{i=1}^{n}\left( \frac{S_i^2+r_i\|h_i\|_{\mathcal{K}_0}^2}{\left(|\dict_{i-1}|+1\right)^2}+ \epsilon_i \right)\left(\prod_{j=i}^{n-1} \frac{|\dict_j|}{|\dict_j|+1}\right)^2 \right]\leq C \frac{n\log(n)}{f(n)^2}.
\end{equation}
Finally we revisit \eqref{eq: initial recursion with r} to achieve our desired conclusion:
\begin{equation}
\label{eq: final thinned bound}
\begin{split}
\mathbb{E}\left[\ksd(q_{\dict_n})^2\right] & \leq \mathbb{E}\left[\exp\left(\sum_{j=1}^n \frac{1}{r_j}\right)\sum_{i=1}^{n} \left( \frac{S_i^2+r_i\|h_i\|_{\mathcal{K}_0}^2}{\left(|\dict_{i-1}|+1\right)^2}+ \epsilon_i \right)\left(\prod_{j=i}^{n-1} \frac{|\dict_j|}{|\dict_j|+1}\right)^2 \right]\\
&= \mathbb{E}\left[e\sum_{i=1}^n\left( \frac{S_i^2+r_i\|h_i\|_{\mathcal{K}_0}^2}{\left(|\dict_{i-1}|+1\right)^2}+ \epsilon_i \right)\left(\prod_{j=i}^{n-1} \frac{|\dict_j|}{|\dict_j|+1}\right)^2\right]\\
&\leq C\frac{n\log(n)}{f(n)^2}\\
\end{split}
\end{equation}
The equality in the second line follows from the fact that $r_j=n$ for all $j$. The final inequality follows from an application of  \eqref{eq: T1 T2 bound substitution} to reach the desired conclusion of Theorem \ref{thm: iid convergence decaying budget}.\hfill $\blacksquare$


\begin{corollary}[i.i.d. Convergence with constant thinning budget]
\label{thm: iid constant budget convergence corollary}
Fix the desired KSD convergence radius $\Delta$. Assume that the dictionary model order growth rate $f$ takes the parametric form $f(i)=\sqrt{i^{1+\alpha}\log(i)}$ with $1< \alpha < 2$. With constant thinning budget $\epsilon=\mathcal{O}\left(\Delta^{1+\frac{1}{\alpha}}\right)$, after $n=\mathcal{O}\left(\frac{1}{\Delta^{\frac{1}{\alpha}}}\right)$ steps the KSD satisfies
\begin{equation}
    \label{eq: iid contant budget corollary conclusion}
    \mathbb{E}\left[\ksd\left(q_{\dict_n}\right)^2\right]\leq C\Delta
\end{equation}
for some generic constant $C$. Equivalently, if we fix a constant thinning budget $\epsilon$, after $n=\mathcal{O}\left(\epsilon^{-\frac{2}{\alpha+1}}\right)$ steps the KSD satisfies 
\begin{equation}
    \label{eq: iid contant budget corollary conclusion alternate}
 \mathbb{E}\left[\ksd\left(q_{\dict_n}\right)^2\right]\leq C\epsilon^{\frac{2}{1+\frac{1}{\alpha}}}
\end{equation}
\end{corollary}

\paragraph{Proof} We will revisit terms $T_1$ and $T_2$ in \eqref{eq: T1 T2 introduction} from the proof of Theorem~\ref{thm: iid convergence decaying budget} and show how to obtain the desired convergence rate in the constant thinning budget regime. 

First, \eqref{eq: final T1 bound} demonstrates that the sampling error term $T_1$ converges at a rate of $\frac{n\log(n)}{f(n)^2}$ (up to generic constants). We set the convergence rate equal to the convergence radius and apply algebraic manipulations to derive the number of sampling and thinning iterations required. In particular, write:
\begin{equation}
    \label{eq: algebraic manipulations constant budget}
    \Delta=\frac{n\log(n)}{f(n)^2}=\frac{n\log(n)}{n^{1+\alpha}\log(n)}=\frac{1}{n^{\alpha}} 
\end{equation}
Since the convergence rate $\frac{n\log(n)}{f(n)^2}$ holds up to generic constants we can conclude that after $n= \mathcal{O}\left(\frac{1}{\Delta^{\frac{1}{\alpha}}}\right)$ steps $T_1\leq C\Delta$.

Next we bound $T_2$, the error associated with memory-reduction. We demonstrated in the proof of Theorem \ref{thm: iid convergence decaying budget} that 
\begin{equation}
T_2\leq \frac{1}{n^2}\sum_{i=1}^{n} \epsilon_i i^2.
\end{equation}
Since $\epsilon_i=\epsilon=\mathcal{O}\left(\Delta^{1+\frac{1}{\alpha}}\right)$ is constant with respect to the summation index $i$, we can evaluate the expression above as 
\begin{equation}
\begin{split}
    \frac{1}{n^2}\sum_{i=1}^{n} \epsilon_i i^2&=\frac{\Delta^{1+\frac{1}{\alpha}}}{n^2}\sum_{i=1}^n i^2\\
    &\leq\frac{\Delta^{1+\frac{1}{\alpha}}}{n^2}Cn^3\\
    &=Cn\Delta^{1+\frac{1}{\alpha}}\\
    &=C\frac{\Delta^{1+\frac{1}{\alpha}}}{\Delta^{\frac{1}{\alpha}}}=C\Delta
\end{split}
\end{equation}
In the first line we plug in $\epsilon$. In the second line we make use of the general summation formula $\sum_{i=1}^n i^2=\frac{n(n+1)(n+2)}{6}\leq Cn^3$. In the final line we apply the fact that $n= \mathcal{O}\left(\frac{1}{\Delta^{\frac{1}{\alpha}}}\right)$ and conclude that $T_2\leq C\Delta$ for some generic constant $C$.

Therefore we have $T_1+T_2\leq C\Delta$ which is the desired conclusion. \hfill $\blacksquare$

Next we consider the case when samples are generated using any MCMC sampling procedure. The one constraint we place on the sampler is that it exhibits geometric convergence with respect to a given function $V$. The function $V$, which controls the rate of geometric convergence, will contribute to the sampling errror term  $T_1$ in our analysis.
\begin{definition}
\label{def: vue definition}
A Markov chain $\{\mat{y}_n\}\subset \mathcal{X}$ with $n$-th step transition kernel $P^n$ is {\bf $V$-uniformly ergodic} for a positive function $V:\mathcal{X}\rightarrow [1,\infty]$ if there exists $R\in [0,\infty],\rho\in (0,1)$ such that 
\begin{equation}
    \label{eq: vue definition}
    \sup_{\mat{x}\in X} \frac{\|P^n(\mat{y},\mat{x})-p(\mat{x})\|}{V(\mat{x})}\leq RV(\mat{y})p^n
\end{equation}
\end{definition}

We introduce new notation from \citep{supchen2019stein} by defining the function $V_{\pm}$:
\begin{equation}
    \label{eq: V+- definition}
    V_{\pm}(s) = \sup_{\mat{x}:k_0(\mat{x},\mat{x})\leq s^2} k_0(\mat{x},\mat{x})^{1/2}V(x)^{\pm 1}
\end{equation}
We will use these two functions $V_{\pm}$ and a result from \cite{supchen2019stein} to control the sampling error of a $V$-uniformly ergodic Markov chain. First we extend the convergence guarantees of Algorithm \ref{alg: outer loop} to general $V$-uniformly ergodic samplers in Theorem \ref{thm: v-u ergodic convergence decaying budget}. Then, when MALA is $V$-uniformly ergodic, the desired result follows.

\begin{theorem}[V-Uniformly Ergodic MCMC Sampling and Decaying Budget]
\label{thm: v-u ergodic convergence decaying budget}
Suppose the same conditions as Lemma ~\ref{lem: thinned spmcmc recursion lemma}, and the Stein kernel $k_0$ satisfies $\mathbb{E}_{\mat{y}\sim P}\left[e^{\gamma k_0(\mat{y},\mat{y})}\right]=b < \infty$, and the lower-bound on the dictionary growth condition $|\dict_i|\geq Cf(i)$  holds where $f(i)= \Omega\left(\sqrt{i\log(i)}\right)$, with compression budgets set as $\epsilon_i=\frac{\log(i)}{f(i)^2}$. Further assume that the candidate samples in each $\{\mat{y}_{i,l}^{m_i}\}$ are produced by a V-uniformly ergodic markov chain. Then the iterate $\dict_{n}=KSDT(\dict_{n-1}\cup \{\mat{x}_n\},\epsilon_n)$ satisfies
\begin{equation}
\label{eq: vue convergence decaying budget conclusion}
\mathbb{E}\left[\ksd(q_{\dict_n})^2 \right]\leq C \left(\frac{n\log(n)}{f(n)^2}+\frac{1}{nf(n)^2}\sum_{i=1}^n V_+(S_i)V_-(S_i)\right)
\end{equation}
\end{theorem}

\paragraph{Proof} This proof is similar to the proof of Theorem \ref{thm: iid convergence decaying budget}. It differs only in the bound applied to the term $\|h_j\|_{\mathcal{K}_0}^2$ which is a subcomponent of the sampling bias term $T_1$. This difference is due to the non-i.i.d. sampling procedure used to generate candidate samples that in turn define $h_j$. When the candidate samples $\{\mat{y}_{i,l}\}^{m_i}_{l=1}$ are generated from a $V$-uniformly ergodic markov chain, we can apply the following bound from \cite{supchen2019stein}[Appendix A.2, Equation (22)]:
\begin{equation}
    \label{eq: h bound vue}
    \mathbb{E}\left[\|h_j\|_{\mathcal{K}_0}^2 \right]\leq \frac{4b}{\gamma}\exp\left({-\frac{\gamma}{2}S_j^2}\right)+2RV_+(S_j)V_-(S_j)\left(\frac{1+2p}{1-p}\right)\frac{1}{m_j}
\end{equation}
where the constants $p$ and $R$ come from the $V$-uniform ergodicity of the Markov chain that generates the candidate samples.

Our goal is to mimic the conclusion of \eqref{eq: apply h bound iid decaying budget} in our current setting.

\begin{align}
\label{eq: apply h bound vue decaying budget}
   \mathbb{E}\Bigg[\sum_{i=1}^{n} &\left( \frac{S_i^2+r_i\|h_i\|_{\mathcal{K}_0}^2}{\left(|\dict_{i-1}|+1\right)^2}\right)\left(\prod_{j=i}^{n-1} \frac{|\dict_j|}{|\dict_j|+1}\right)^2 \Bigg] \nonumber\\
    &\leq C\frac{1}{f(n)^2} \sum_{i=1}^{n} \left(S_i^2+r_i\exp\left(-\frac{\gamma}{2}S_i^2\right) +\frac{V_+(S_i)V_-(S_i)}{m_j}\right)
\end{align}
The first inequality is an application of \eqref{eq: h bound vue} to the final conclusion of \eqref{eq: bound computation} from the proof of Theorem \ref{thm: iid convergence decaying budget}. We also absorb the constant factors into $C$. We select $S_i=\sqrt{\frac{2}{\gamma}\log(n\wedge m_i)}$, $r_i=n$, and $m_i=n$ to achieve the bound
\begin{align}
    \mathbb{E}\Bigg[\sum_{i=1}^{n} \left( \frac{S_i^2+r_i\|h_i\|_{\mathcal{K}_0}^2}{\left(|\dict_{i-1}|+1\right)^2}\right)&\left(\prod_{j=i}^{n-1} \frac{|\dict_j|}{|\dict_j|+1}\right)^2\Bigg]\nonumber\\
    &\leq C\frac{1}{f(n)^2}\sum_{i=1}^{n} \left(\log(n)+\frac{V_+(S_i)V_-(S_i)}{n}\right) \nonumber\\
    &\leq C\left(\frac{n\log(n)}{f(n)^2}+\frac{1}{nf(n)^2}\sum_{i=1}^n V_+(S_i)V_-(S_i)\right)
    \label{eq: final T1 bound vue}
\end{align}
The first inequality is the conclusion of the previous equation with $S_i$, and $r_i$ evaluated and all constants absorbed. The second inequality holds because the summand does not depend on the index so we can multiply by $n$. This completes the bound on $T_1$ in the setting where samples are generated by a V-uniformly ergodic Markov chain.

The bound on $T_2$ in Theorem \ref{thm: iid convergence decaying budget} does not depend on the sample generation mechanism, so we can draw the same conclusion as \eqref{eq: final T2 bound} which does not introduce any terms with higher asymptotic growth order than $n\log(n)/f(n)^2$. Therefore the bound from \eqref{eq: final T1 bound vue} is the asymptotic upper bound, and we have achieved the desired result. \hfill $\blacksquare$

Our goal is to show that using MALA to generate samples ensures convergence of Algorithm \ref{alg: outer loop}. When MALA is inwardly convergent, MALA is $V$-uniformly ergodic. We make use of this fact to add one more condition (inwardly convergent) and present a convergence result for Algorithm \ref{alg: outer loop} using the MALA sampler.

\begin{corollary}[MALA Convergence with Decaying Budget]
\label{thm: mala convergence decaying budget}
Assume that the candidate sample sets $\{\mat{y}_{i,l}\}_{l=1}^{m_i}$ are generated by the Metropolis-Adjusted Langevin Algorithm (MALA) transition kernel with stepsize $h$ and that MALA is inwardly convergent for the target density $p$. Further assume that $p$ is distantly dissipative. Then the iterates $\dict_n$ of Algorithm \ref{alg: outer loop} satisfy
\[
\mathbb{E}\left[\ksd(q_{\dict_n})^2\right]\leq C \left(\frac{n\log(n)}{f(n)^2}\right)
\]
\end{corollary}
\paragraph{Proof} We make use of two results from \cite{supchen2019stein}, Appendix A.3. First we use the fact that MALA is inwardly convergent to conclude that MALA is $V$-uniformly ergodic with $V(\mat{x})=1+\|x\|_2$. The second result is the set of bounds $V_+(s)\leq C(s+s^2)$ and $V_-(s)\leq C$ for some generic constant $C$. Then we can plug in $S_i=\sqrt{\frac{2}{\gamma}\log(n)}$ (since $m_i=n$) to bound the second term in Equation \ref{eq: final T1 bound vue}:
\begin{align}
    \frac{1}{nf(n)^2}&\sum_{i=1}^n V_+(S_i)V_-(S_i) \nonumber\\
    &\leq \frac{C}{nf(n)^2}\sum_{i=1}^n \left(S_i+S_i^2\right) \nonumber\\
    &\leq \frac{C}{nf(n)^2}\sum_{i=1}^n \left( \sqrt{\log(n)}+\log(n) \right)\nonumber\\
    &\leq \frac{C\sqrt{log(n)}+\log(n)}{f(n)^2}\nonumber\\
    &\leq \frac{C\log(n)}{f(n)^2}
    \label{eq: MALA v-term bound}
\end{align}

The first inequality is result of the evaluation of $V_\pm(S_i)$ and the absorption of constants. The second inquality is the results of the evaluation of $S_i=\sqrt{\frac{2}{\gamma}\log(n)}$ and the absorption of constants. The third inequality follows from the fact that the summand does not depend on the index. The final inequality is an application of the fact that $\sqrt{\log(n)}\leq\log(n)$ for $n\geq 10$, which we assume as a reasonable minimum number of steps.

Since the final term in \eqref{eq: MALA v-term bound} is asymptotically upper bounded by $n\log(n)/f(n)^2$ due to the extra factor of $n$ we can ignore this term in Equation \ref{eq: vue convergence decaying budget conclusion} and draw the desired conclusion that 
\begin{equation}
\label{eq: vue convergence decaying budget conclusion proof}
\mathbb{E}\left[\ksd(q_{\dict_n})^2 \right]\leq C \left(\frac{n\log(n)}{f(n)^2}\right)
\end{equation}
\hfill $\blacksquare$

\begin{corollary}[MALA Convergence with constant thinning budget]
\label{thm: mala constant budget convergence corollary}
Fix the desired KSD convergence radius $\Delta$. Assume that the dictionary model order growth rate $f$ takes the parametric form $f(i)=\sqrt{i^{1+\alpha}\log(i)}$ with $1< \alpha < 2$. With constant thinning budget $\epsilon=\mathcal{O}\left(\Delta^{1+\frac{1}{\alpha}}\right)$, after $n=\mathcal{O}\left(\frac{1}{\Delta^{\frac{1}{\alpha}}}\right)$ steps the KSD satisfies
\begin{equation}
    \label{eq: mala contant budget corollary conclusion}
    \mathbb{E}\left[\ksd\left(q_{\dict_n}\right)^2\right]\leq c\Delta
\end{equation}
for some generic constant $C$.
Equivalently, if we fix a constant thinning budget $\epsilon$, after $n=\mathcal{O}\left(\epsilon^{-\frac{2}{\alpha+1}}\right)$ steps the KSD satisfies 
\begin{equation}
    \label{eq: mala contant budget corollary conclusion alternate}
    \mathbb{E}\left[\ksd\left(q_{\dict_n}\right)^2\right]\leq C\epsilon^{\frac{2}{1+\frac{1}{\alpha}}}
\end{equation}
\end{corollary}
This proof is exactly the same as the proof of Corollary \ref{thm: iid constant budget convergence corollary} since the asymptotic convergence rate achieved in Corollary \ref{thm: mala convergence decaying budget}  is exactly the same as in the i.i.d. case in Theorem \ref{thm: iid convergence decaying budget}. We note that the convergence rate from Corollary \ref{thm: mala constant budget convergence corollary} also trivially extends to the general case of V-uniformly ergodic samplers using the result of Theorem \ref{thm: v-u ergodic convergence decaying budget} and the analysis of Corollary \ref{thm: constant budget convergence corollary}.

%% file: sections/experiments_appendix.tex
\section{Additional Experiments\label{sec: experiments appendix}}

In the main body we focused on results using the IMQ base kernel for the KSD. In this Appendix section we report results for the RBF kernel and additional metrics for both the IMQ and RBF kernels. We also provide per-sampler figures so that readers can easily evaluate the affect of our approach on individual samplers. We keep the same experimental settings as in the main body of the text. We follow \cite{detommaso2018stein} and set the kernel bandwith $h$ as the problem dimension $h=dim(\mat{x})$ when using the RBF kernel.

\subsection{Goodwin Oscillator}\label{sec: goodwin appendix}
\begin{figure}
    \centering
    \includegraphics[width=\textwidth]{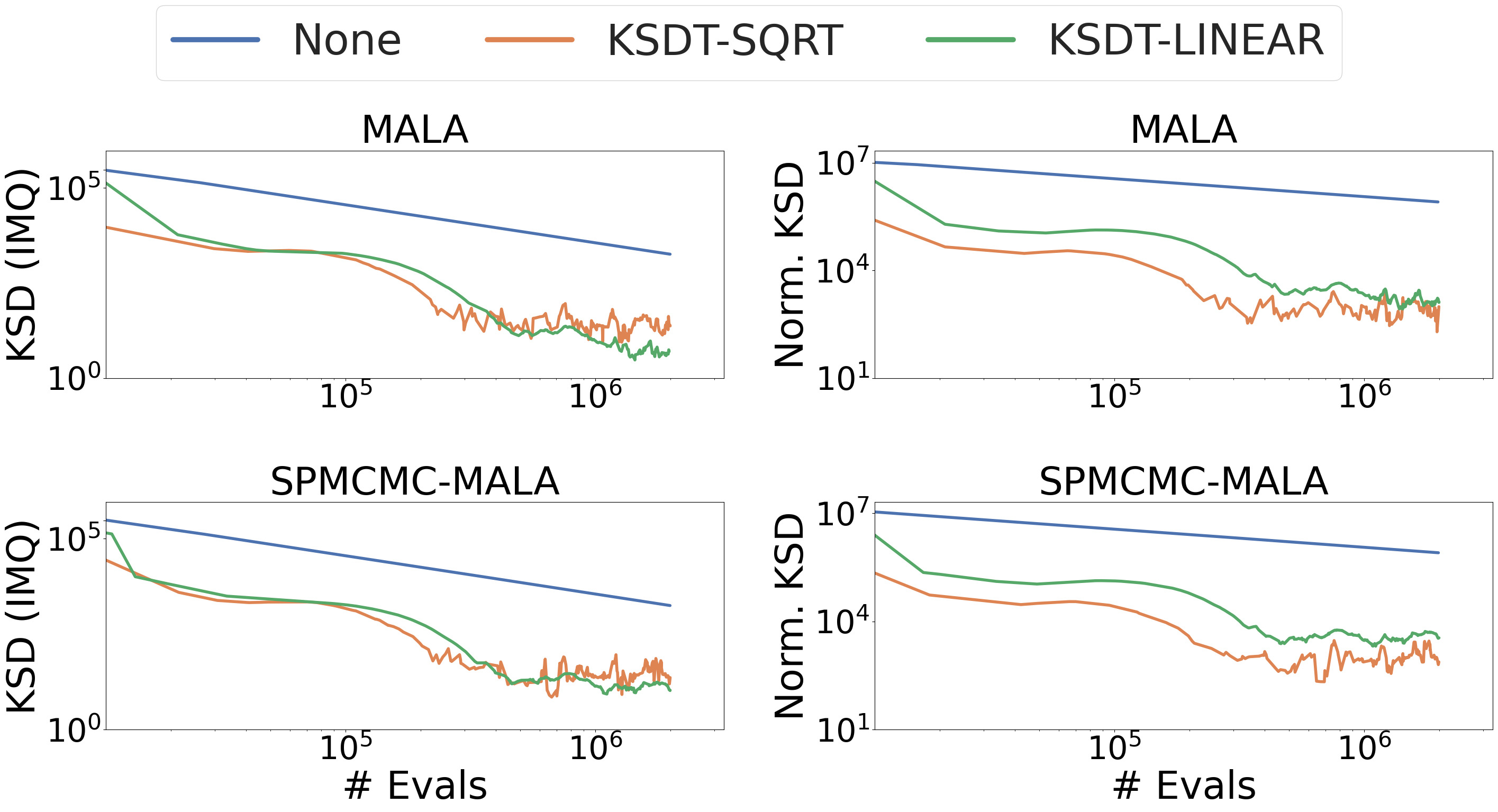}
    \caption{Goodwin problem, IMQ base kernel. Comparison of un-thinned sample chains from the MALA and SPMCMC-MALA samplers with our thinning methods KSDT-SQRT and KSDT-LINEAR. Lower KSD and lower Normalized KSD indicate more accurate representations of the target distribution. Both axes are log-scale. Our methods outperform the baseline methods on both KSD and Normalized KSD.}
    \label{fig: per-sampler goodwin imq mala}
\end{figure}

\begin{figure}
    \centering
    \includegraphics[width=\textwidth]{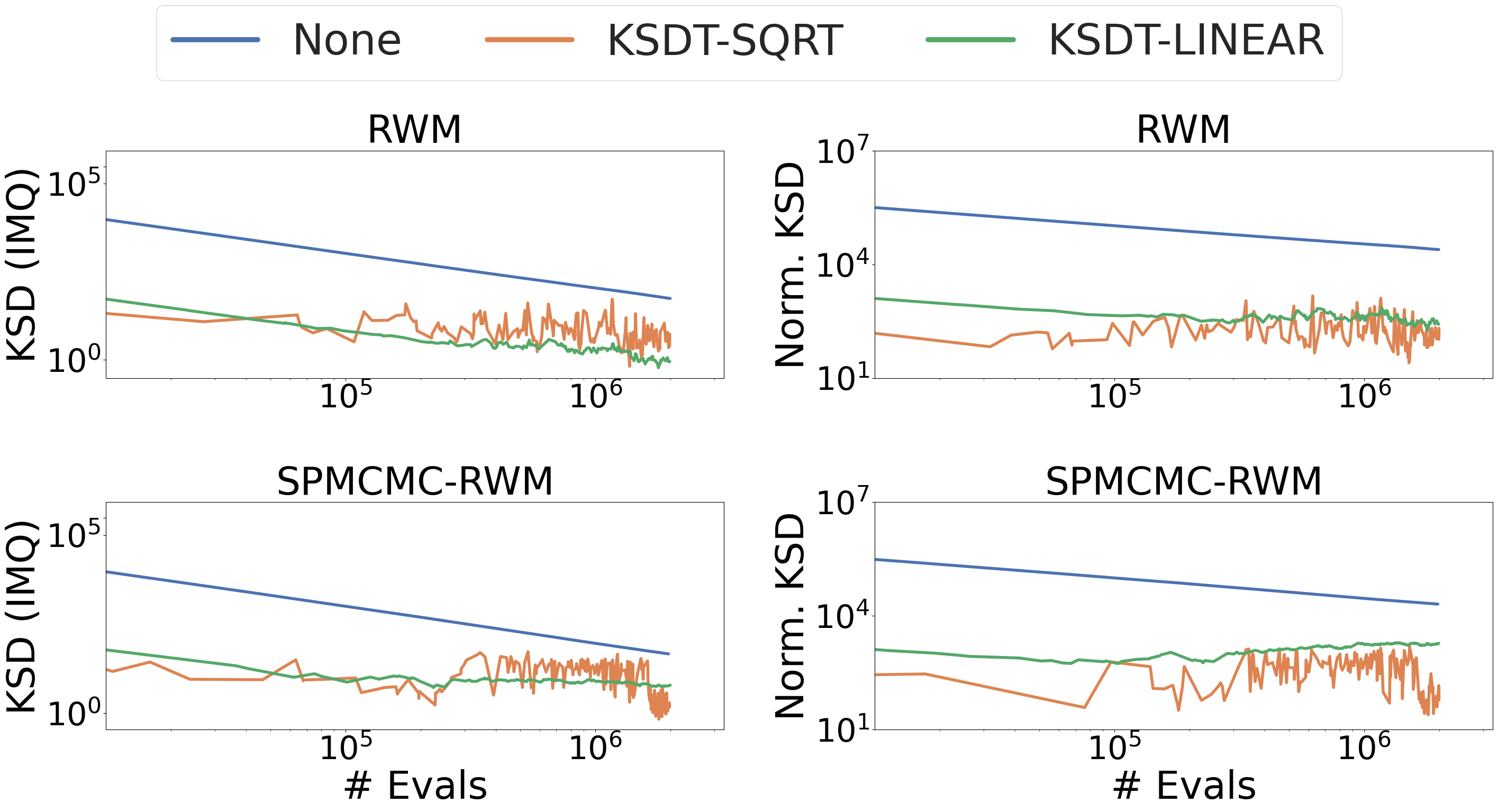}
    \caption{Goodwin problem, IMQ base kernel. Comparison of un-thinned sample chains from the RWM and SPMCMC-RWM samplers with our thinning methods KSDT-SQRT and KSDT-LINEAR. Lower KSD and lower Normalized KSD indicate more accurate representations of the target distribution. Both axes are log-scale. Our methods outperform the baseline methods on both KSD and Normalized KSD.}
    \label{fig: per-sampler goodwin imq rwm}
\end{figure}

\begin{figure}
    \centering
    \includegraphics[width=\textwidth]{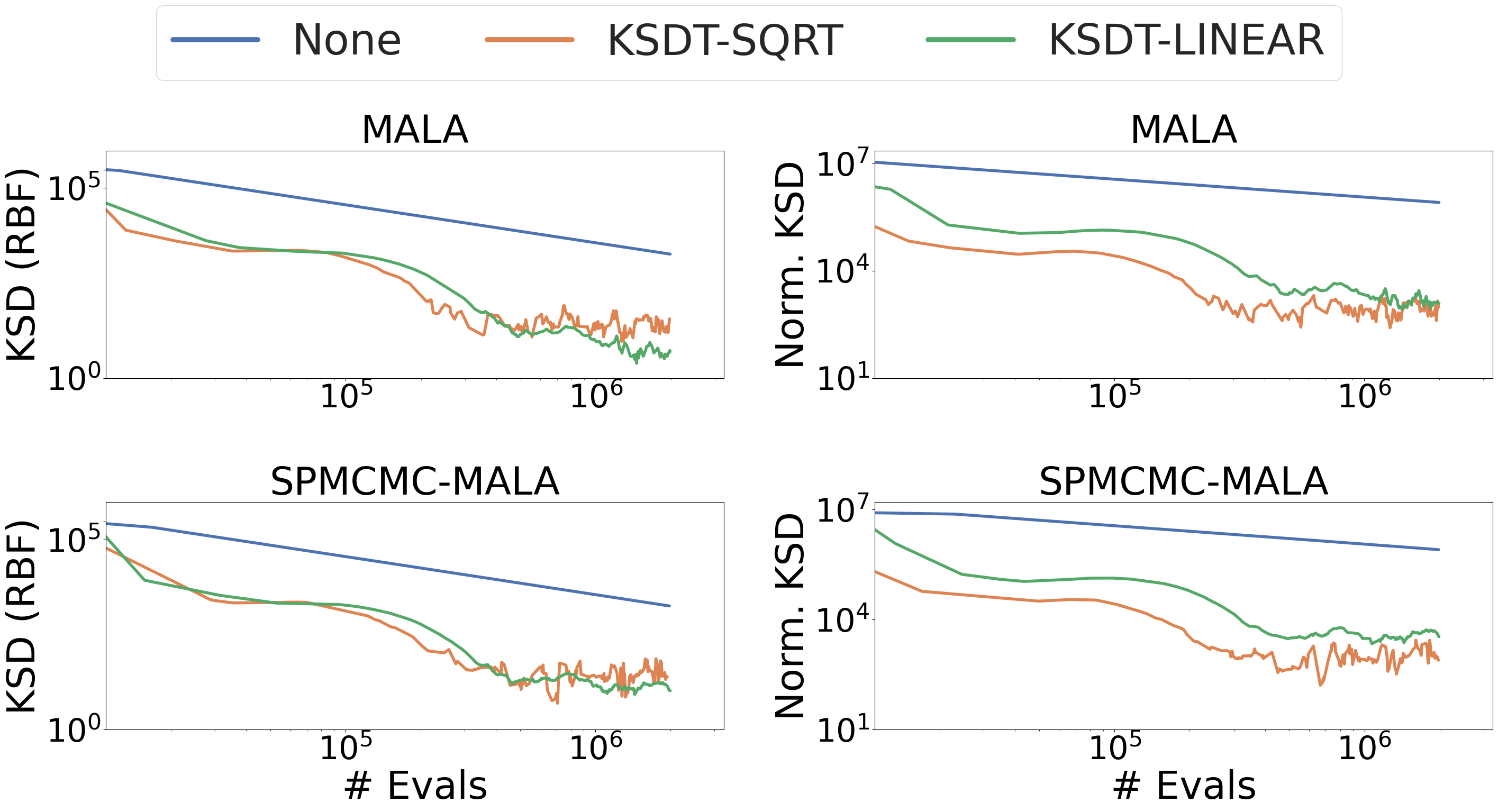}
    \caption{Goodwin problem, RBF base kernel. Comparison of un-thinned sample chains from the MALA and SPMCMC-MALA samplers with our thinning methods KSDT-SQRT and KSDT-LINEAR. Lower KSD and lower Normalized KSD indicate more accurate representations of the target distribution. Both axes are log-scale. Our methods outperform the baseline methods on both KSD and Normalized KSD.}
    \label{fig: per-sampler goodwin rbf mala}
\end{figure}

\begin{figure}
    \centering
    \includegraphics[width=\textwidth]{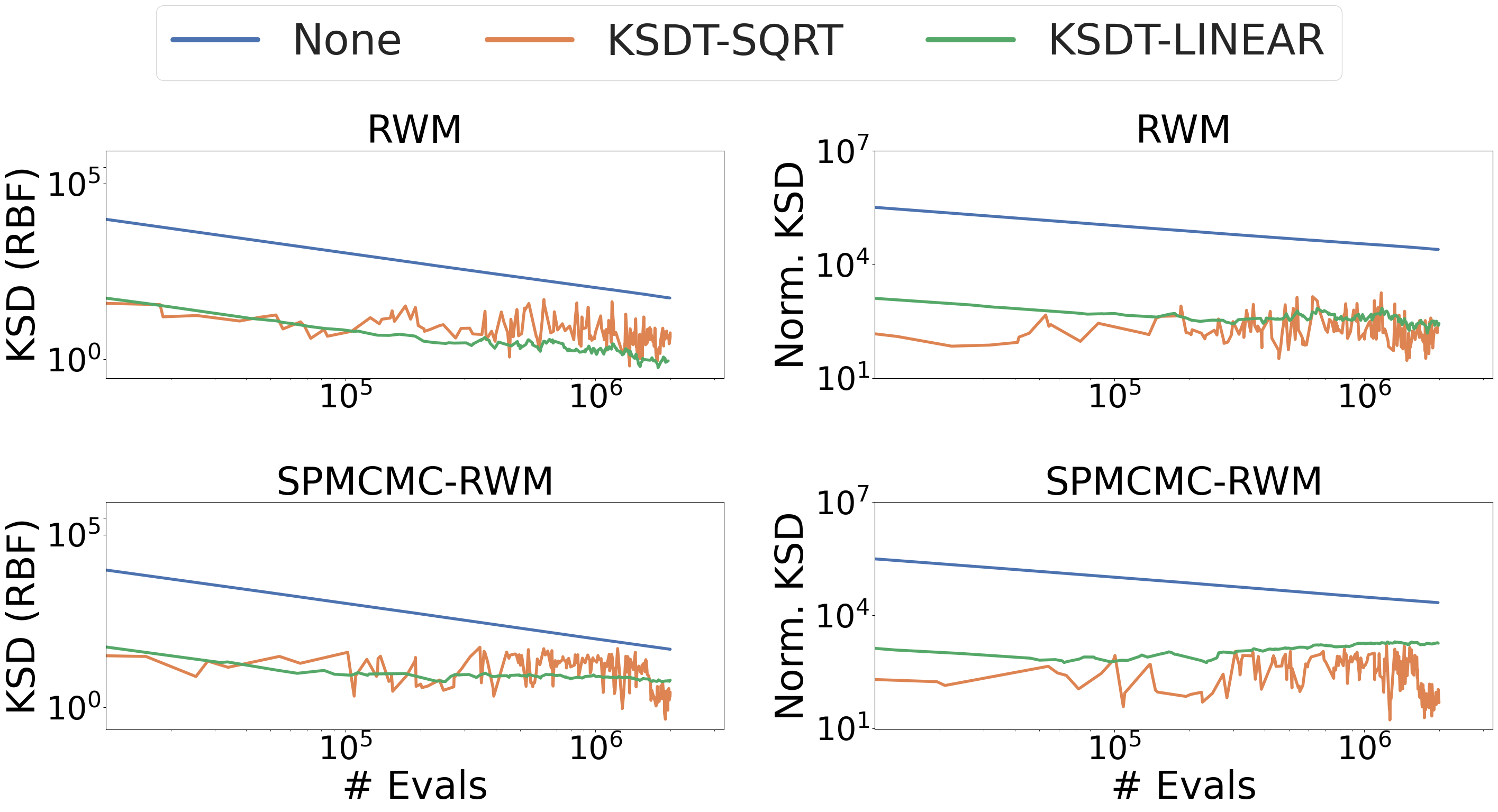}
    \caption{Goodwin problem, RBF base kernel. Comparison of un-thinned sample chains from the RWM and SPMCMC-RWM samplers with our thinning methods KSDT-SQRT and KSDT-LINEAR. Lower KSD and lower Normalized KSD indicate more accurate representations of the target distribution. Both axes are log-scale. Our methods outperform the baseline methods on both KSD and Normalized KSD.}
    \label{fig: per-sampler goodwin rbf rwm}
\end{figure}

We report the results on a per-sampler basis when applying KSD thinning to both the RWM chain and the MALA chains with and without the SPMCMC update rule using either the RBF or IMQ kernels in Figures \ref{fig: per-sampler goodwin imq mala},\ref{fig: per-sampler goodwin rbf mala},\ref{fig: per-sampler goodwin imq rwm}, and \ref{fig: per-sampler goodwin rbf rwm}. In all cases the conclusions are similar to those we draw from Figure \ref{fig: goodwin imq} in the main body. Removing samples using our KSDT-LINEAR and KSDT-SQRT methods improves both KSD and normalized KSD.

\subsection{Calcium Signalling Model}\label{sec: cardiac appendix}
\begin{figure}
    \centering
    \includegraphics[width=\textwidth]{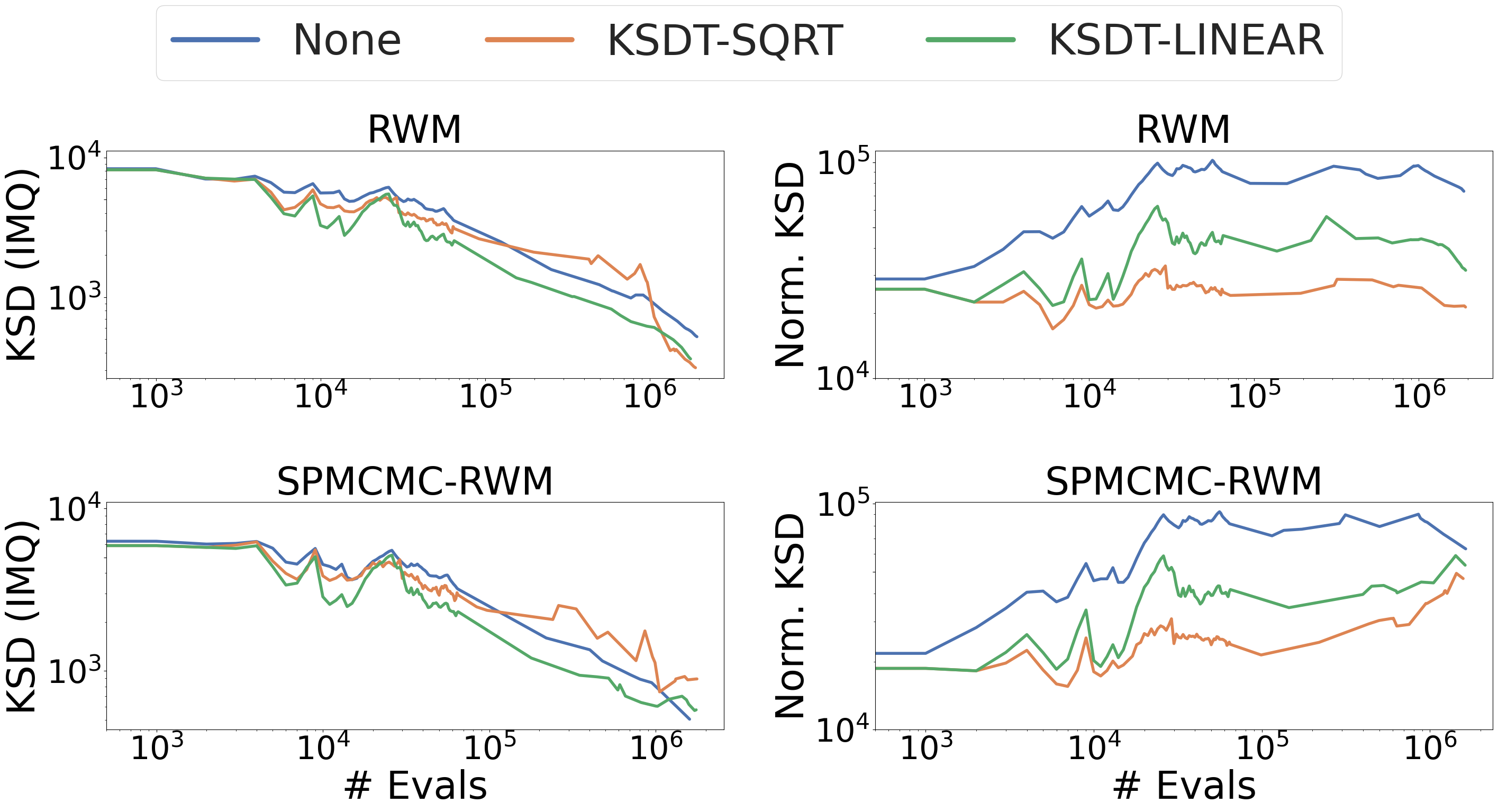}
    \caption{Cardiac problem, IMQ base kernel. Comparison of un-thinned sample chains from the tempered RWM and SPMCMC-RWM samplers with our thinning methods KSDT-SQRT and KSDT-LINEAR. Lower KSD and lower Normalized KSD indicate more accurate representations of the target distribution. Both axes are log-scale. Our methods are competitive on KSD and outperform the baseline methods on Normalized KSD.}
    \label{fig: per-sampler cardiac imq}
\end{figure}

\begin{figure}
    \centering
    \includegraphics[width=\textwidth]{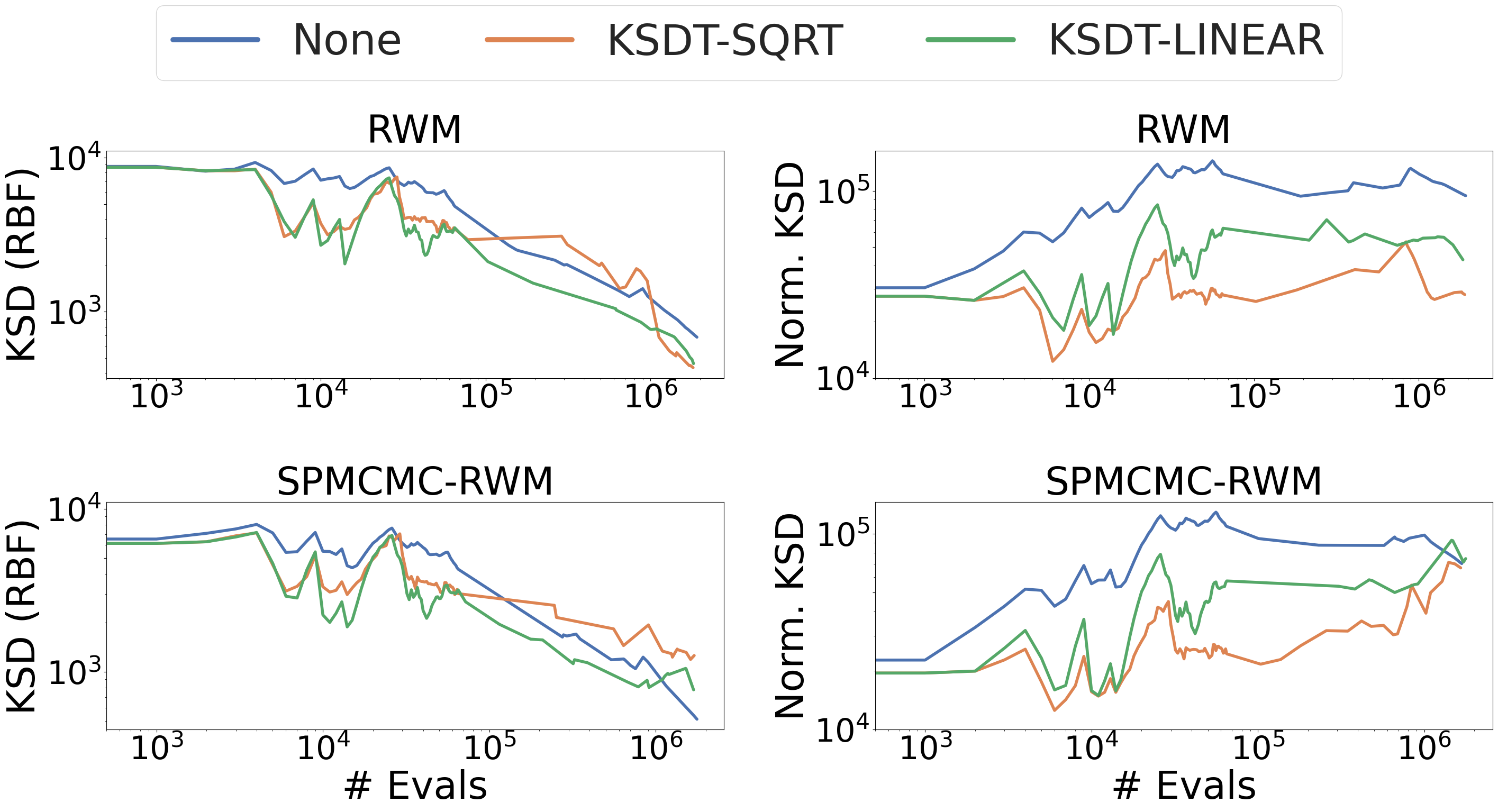}
    \caption{Cardiac problem, RBF base kernel. Comparison of un-thinned sample chains from the tempered RWM and SPMCMC-RWM samplers with our thinning methods KSDT-SQRT and KSDT-LINEAR. Lower KSD and lower Normalized KSD indicate more accurate representations of the target distribution. Both axes are log-scale. Our methods are competitive on KSD and outperform the baseline methods on Normalized KSD.}
    \label{fig: per-sampler cardiac rbf}
\end{figure}

We report the results on a per-sampler basis when applying KSD thinning to the tempered RWM sampler with and without the SPMCMC update rule using either the RBF or IMQ kernels in Figures \ref{fig: per-sampler cardiac imq},\ref{fig: per-sampler cardiac rbf}. In all cases the conclusions are similar to those we draw from Figure \ref{fig: cardiac imq} in the main body. After accounting for model complexity (number of retained samples) with the Normalized KSD metric our methods outperform the baseline samplers. Our KSDT-LINEAR and KSDT-SQRT methods outperforms the baseline sampler more often than not on the (un-normalized) KSD metric as well. 

\subsection{Bayesian Neural Network Subspace Sampling}\label{sec: bnn subspace appendix}

In this appendix we provide additional results on accuracy and calibration as well as results using the RBF base kernel. No batch normalization or data augmentation is used as they do not admit Bayesian interpretations \citep{wenzel2020good}. In all experiments the hyper-parameters were optimized only for the ground truth MCMC chain. We set the SPMCMC parameter to $m=5$ and all methods perform 2000 SGLD steps. We measure the agreement, total variation, and the number of dictionary samples at steps $\{100,200,500,1000,2000\}$ for each task.

\paragraph{Curve Subspace} In order to construct the curve subspace we follow the procedure of \cite{supizmailov2020subspace}. First we pretrain two neural networks with Stochastic Weight Averaging \cite{izmailov2018averaging}. Next, given the weights $\mat{w}_1,\mat{w}_2$ of two pretrained networks we initialize the curve midpoint $\mat{w}_{1/2}=\left(\mat{w}_1+\mat{w}_2\right)$ and define the piece-wise linear curve 
\begin{equation}
    \mat{w}_t = \begin{cases} \mat{w}_1+\frac{t}{0.5}\left(\mat{w}_{1/2}-\mat{w}_1\right) &\text{ if } 0\leq t \leq 0.5\\
     \mat{w}_{1/2}+\frac{t-0.5}{0.5}\left(\mat{w}_2-\mat{w}_{1/2}\right) &\text{ if } 0.5< t \leq 1.0
    \end{cases}
\end{equation} where $t\in\left(0,1\right)$. To train the curve network, for each batch we sample $t\in (0,1)$ and backpropagate gradients only to $\mat{w}_{1/2}$. Finally, we define $\hat{\mat{w}}=(\mat{w}_0+\mat{w}_1)/2$ as the ``base point" and $\mat{v}_1=\mat{w}_0-\hat{\mat{w}}$ and $\mat{v}_2=\mat{w}_{1/2}-\hat{\mat{w}}$ as the subspace vectors. We perform sampling in the 2D subspace centered at $\hat{\mat{w}}$ and spanned by the vectors $\mat{v}_1,\mat{v}_2$.

\begin{figure}
    \centering
    \includegraphics[width=\textwidth]{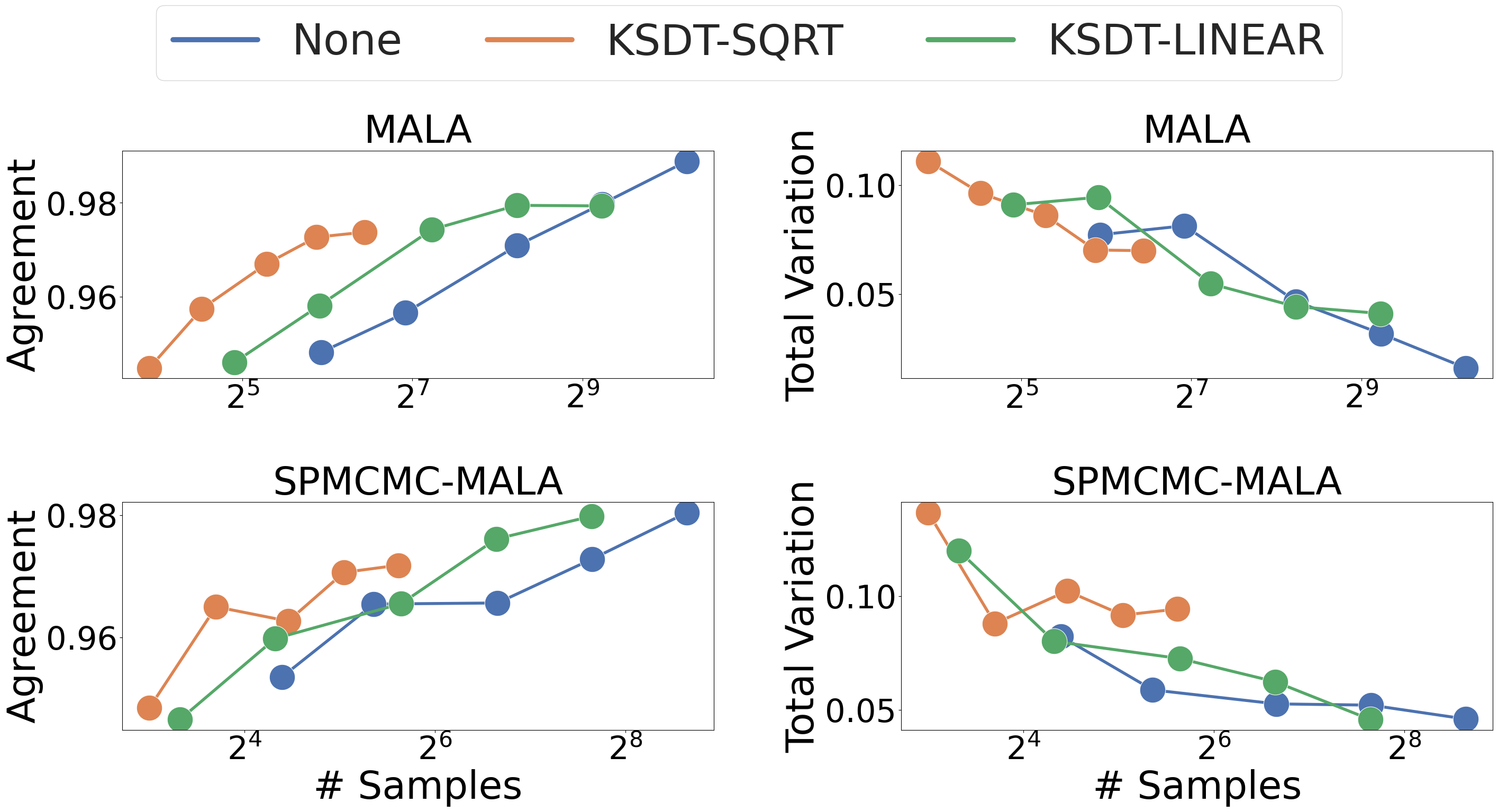}
    \caption{Agreement (left) and total variation (right) with the ground truth CIFAR-10 posterior predictive distribution. Higher Agreement and lower Total Variation indicate more accurate representation of the posterior predictive distribution. Color indicates thinning method (or baseline without thinning). The x-axis is log-scale. RBF base kernel. Our methods Pareto-dominate the baseline samplers on the Agreement metric and improve the Total Variation Pareto frontier in the low-sample regime.} 
    \label{fig: cifar rbf agreement tv}
\end{figure}

\begin{figure}
    \centering
    \includegraphics[width=\textwidth]{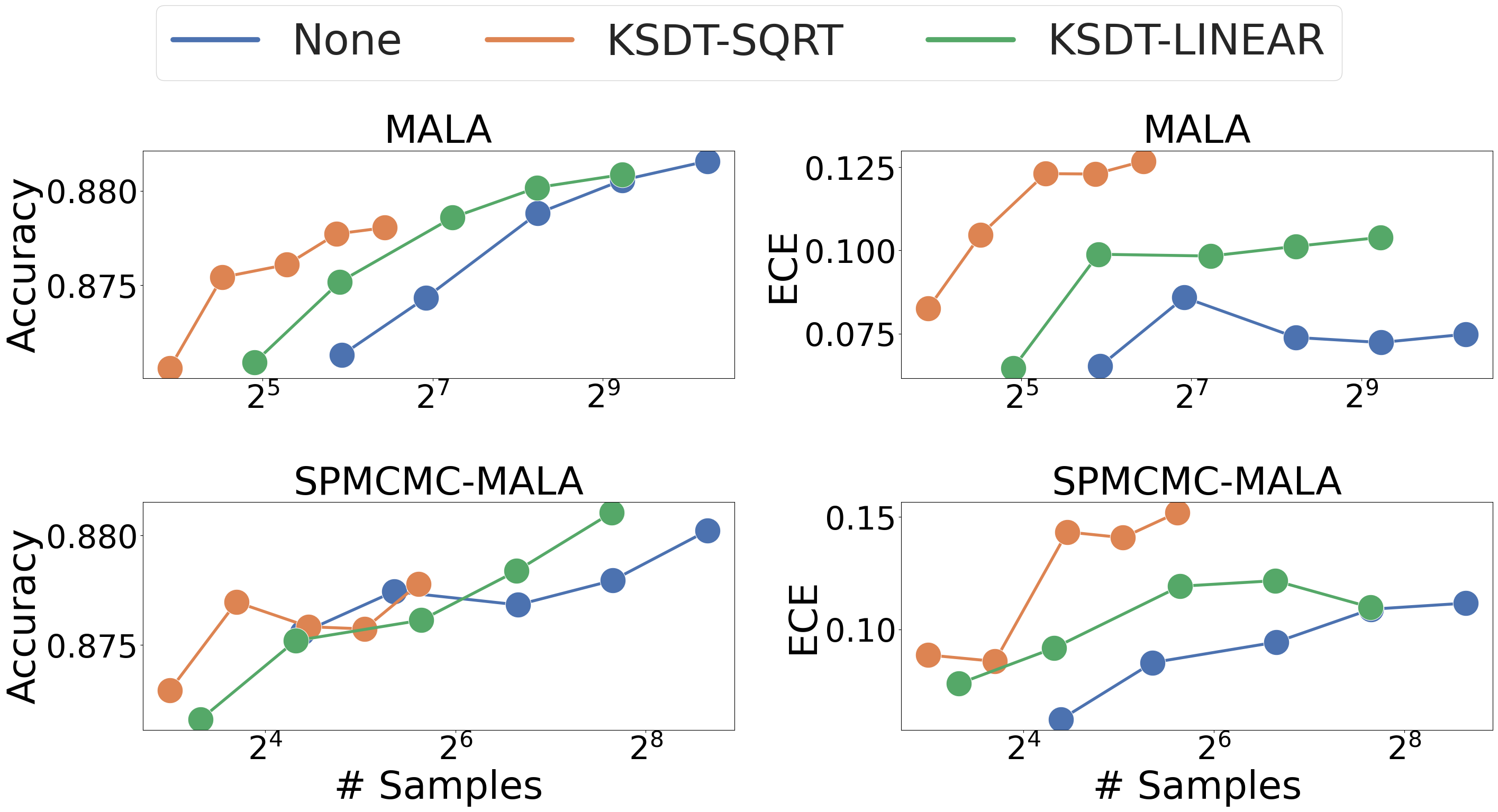}
    \caption{Accuracy (left) and Expected Calibration Error (right) on CIFAR-10 using the RBF base kernel. Higher Accuracy and lower ECE are better. Color indicates thinning method (or baseline without thinning). The x-axis is log-scale. Our methods Pareto-dominate the baseline samplers on accuracy and expand the low-sample Pareto frontier for ECE.} 
    \label{fig: cifar rbf ece acc}
\end{figure}

\begin{figure}
    \centering
    \includegraphics[width=\textwidth]{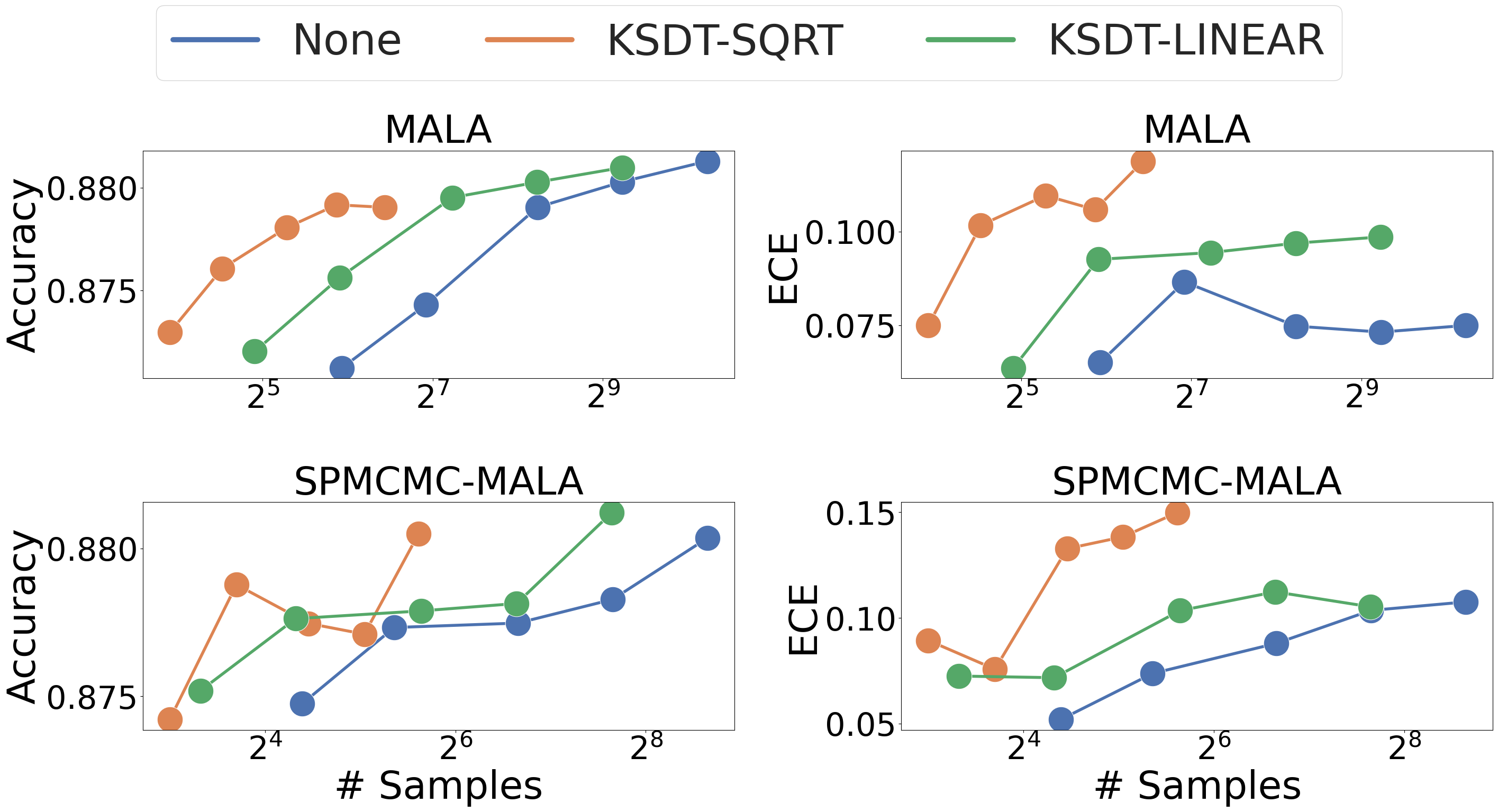}
    \caption{Accuracy (left) and Expected Calibration Error (right) on CIFAR-10 using the IMQ base kernel. Higher Accuracy and lower ECE are better. Color indicates thinning method (or baseline without thinning). The x-axis is log-scale. Our methods Pareto-dominate the baseline samplers on accuracy and expand the low-sample Pareto frontier for ECE.} 
    \label{fig: cifar imq ece acc}
\end{figure}

\paragraph{CIFAR-10 Classification}
We present results for both total variation and ECE when using the RBF kernel for all KSD-based methods in Figure \ref{fig: cifar rbf agreement tv} demonstrate that our online thinning method outperforms both the baseline sampler and SPMCMC-based samplers on the agreement metric. We draw the same conclusion as the main body: improvement is mixed on the total variation metric, where our methods outperform existing methods in the low-sample regime, but have similar performance to existing methods once the model complexity is high. In Figures \ref{fig: cifar rbf ece acc} and \ref{fig: cifar imq ece acc} we report results for accuracy and expected calibration error (ECE) using the RBF and IMQ kernels respectively. We observe that our thinning methods usually Pareto dominate the baseline methods on accuracy, and offer new points on the ECE Pareto frontier. We note that the goal of all methods is to match the predictive posterior distribution, not necessarily to achieve low accuracy. If the ground-truth predictive posterior has low accuracy and high ECE, then methods that perform best on the task of matching the posterior may have lower accuracy and higher ECE.

\begin{figure}
    \centering
    \includegraphics[width=\textwidth]{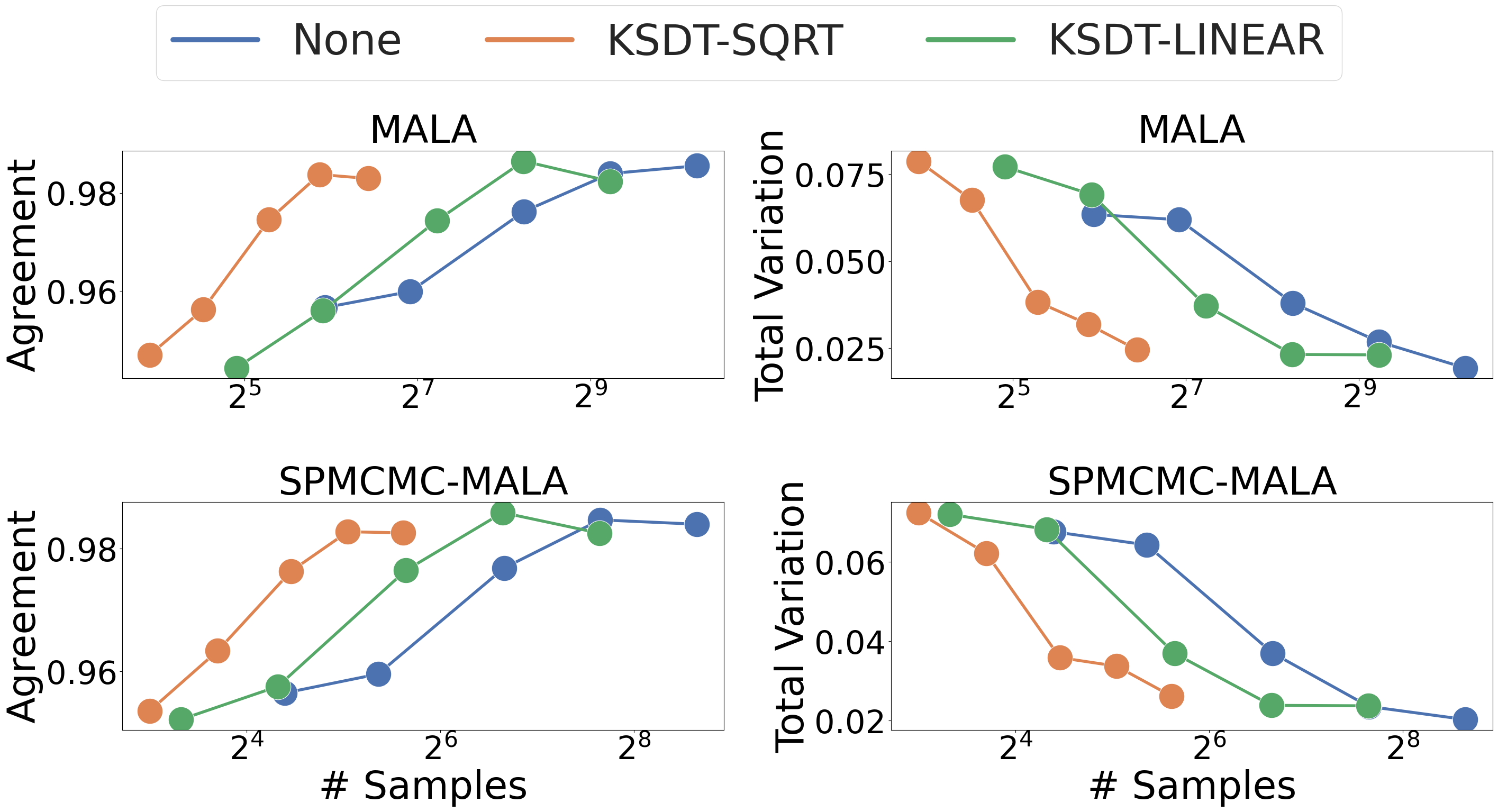}
    \caption{Agreement (left) and total variation (right) with the ground truth IMDB posterior predictive distribution. Higher Agreement and lower Total Variation indicate more accurate representation of the posterior predictive distribution. Color indicates thinning method (or baseline without thinning). The x-axis is log-scale. RBF base kernel. Our methods Pareto-dominate the baseline samplers on both the Agreement and Total Variation metrics.} 
    \label{fig: imdb rbf agreement tv}
\end{figure}

\begin{figure}
    \centering
    \includegraphics[width=\textwidth]{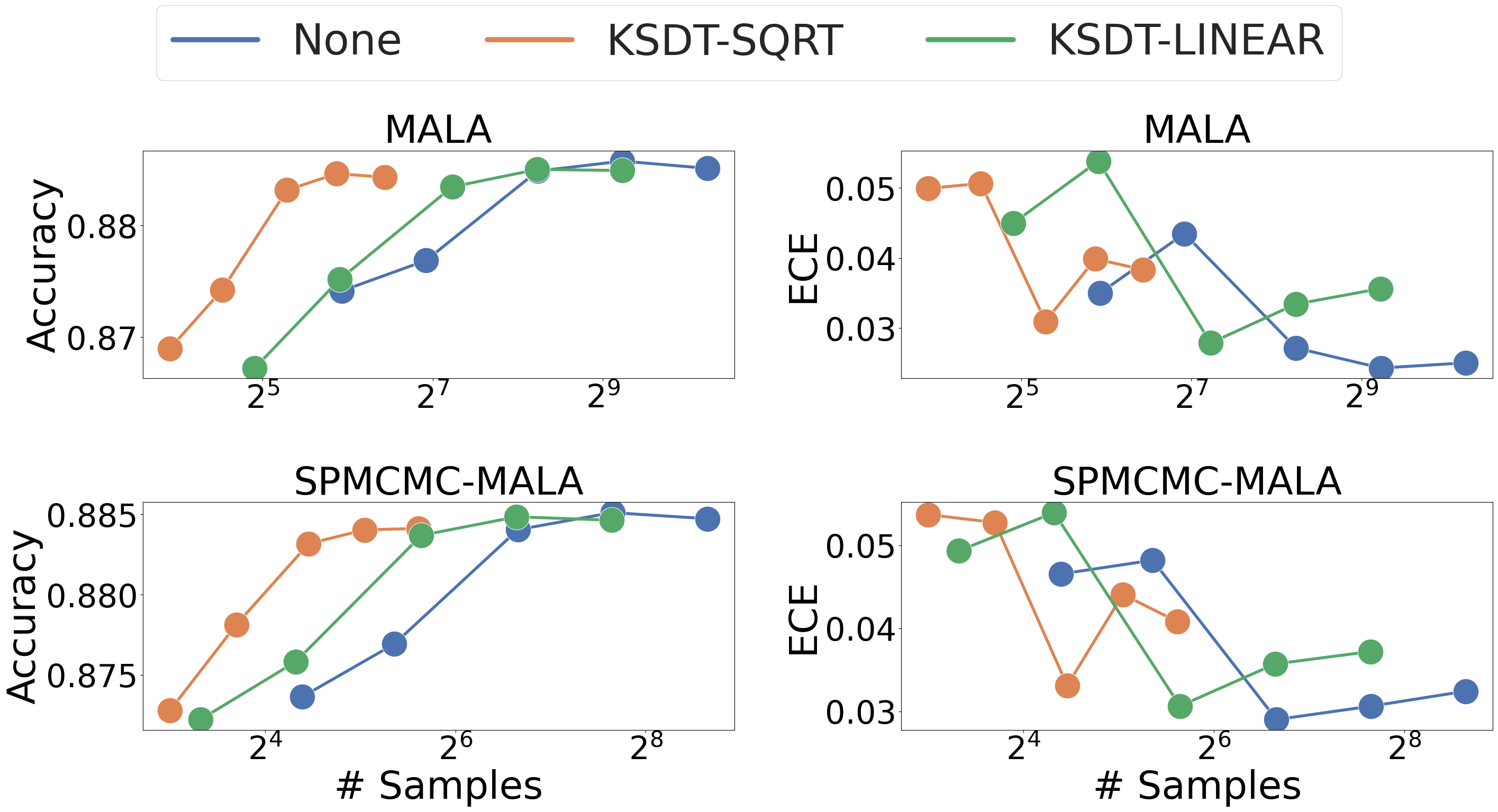}
    \caption{Accuracy (left) and Expected Calibration Error (right) on IMDB using the RBF base kernel. Higher Accuracy and lower ECE are better. Color indicates thinning method (or baseline without thinning). The x-axis is log-scale. Our methods Pareto-dominate the baseline samplers on accuracy and expand the low-sample Pareto frontier for ECE.} 
    \label{fig: imdb rbf ece acc}
\end{figure}

\begin{figure}
    \centering
    \includegraphics[width=\textwidth]{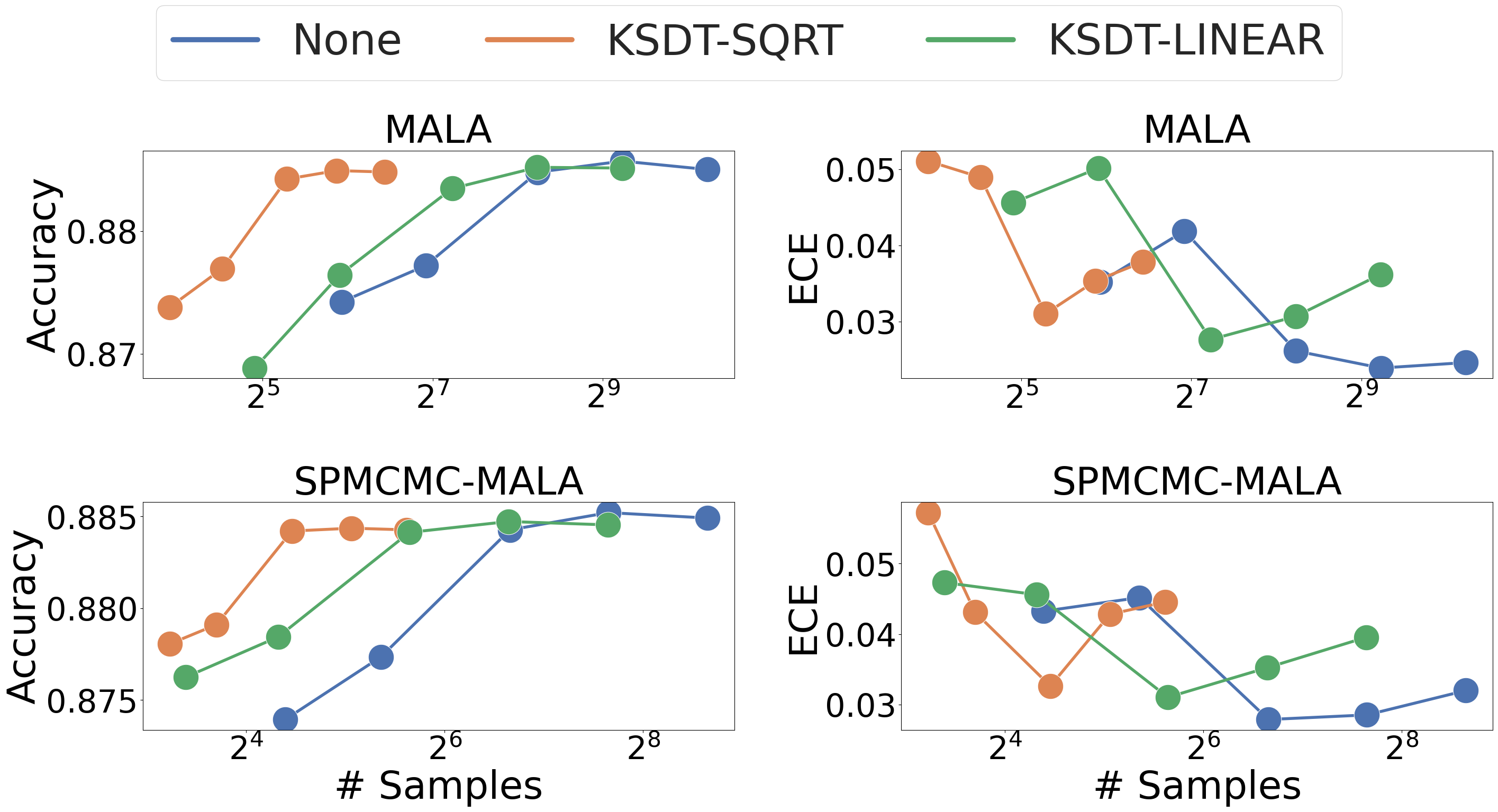}
    \caption{Accuracy (left) and Expected Calibration Error (right) on IMDB using the IMQ base kernel. Higher Accuracy and lower ECE are better. Color indicates thinning method (or baseline without thinning). The x-axis is log-scale. Our methods Pareto-dominate the baseline samplers on accuracy and expand the low-sample Pareto frontier for ECE.} 
    \label{fig: imdb imq ece acc}
\end{figure}

\paragraph{IMDB Sentiment Prediction}
We report results on both total variation and ECE when using the RBF kernel for all KSD-based methods in Figure \ref{fig: imdb rbf agreement tv}. These results imply the same conclusion as the main body text IMQ kernel experiments: our online thinning method outperforms both the baseline sampler and SPMCMC-based samplers on the agreement metric. In Figures \ref{fig: imdb rbf ece acc} and \ref{fig: imdb imq ece acc} we report results for accuracy and expected calibration error (ECE) using the RBF and IMQ kernels respectively. We observe that our thinning methods Pareto dominate the baseline methods on accuracy, and offer new points on the ECE Pareto frontier.

\subsection{Parameter Sensitivity}\label{sec: parameter sensitivity appendix}
\begin{figure}
    \centering
    \includegraphics[width=\textwidth]{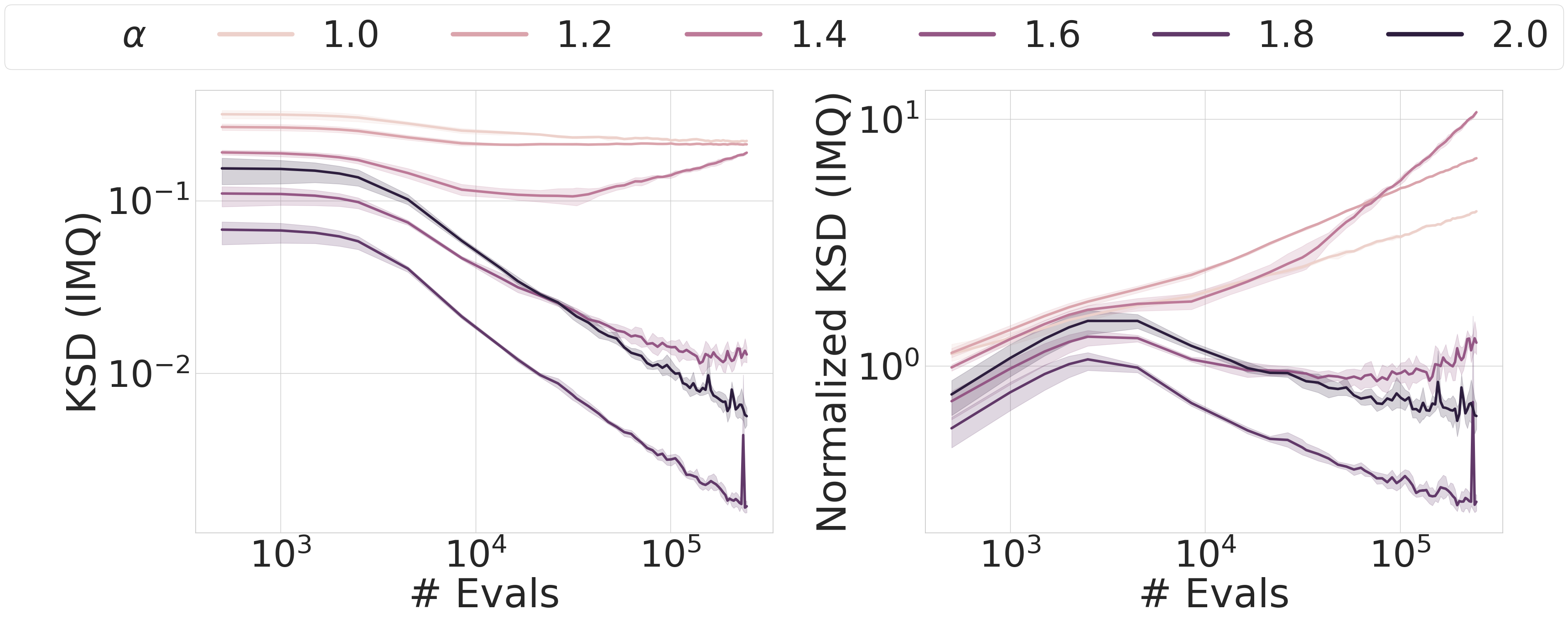}
    \caption{Sensitivity to dictionary growth rate parameter $\alpha$ on a bi-modal gaussian mixture problem. Lower KSD and lower Normalized KSD indicate more accurate representations of the target distribution. These results indicate that tuning the dictionary growth rate has potential to further improve our method.}
    \label{fig: alpha sensitivity}
\end{figure}
We study the sensitivity of our thinning algorithm to the dictionary growth rate $f(i)\in o\left(\sqrt{i\log(i)}\right)$. We target an equally weighted bimodal Gaussian mixture distribution with means $\{(0,0),(1,1)\}$ and covariance $0.5*I_2$ for each mode. To decouple the sampler performance from our thinning algorithm we match the i.i.d. sampling setting of Theorem \ref{thm: decaying budget}. We draw samples directly from the true distribution and apply the SPMCMC update rule with $m=5$. In Figure \ref{fig: alpha sensitivity} we vary the exponent $\alpha$ with budget $f(t)=\sqrt{t^\alpha\log t}/2$. When $\alpha=2.0$ we use linear growth rate $f(t)=t/2$. We observe the results are sensitive to parameter tuning on this toy problem, and that the best setting for both KSD and normalized KSD is $\alpha=1.8$. Similar results for the RBF kernel are presented in Appendix Figure \ref{fig: alpha sensitivity rbf}.
\begin{figure}
    \centering
    \includegraphics[width=\textwidth]{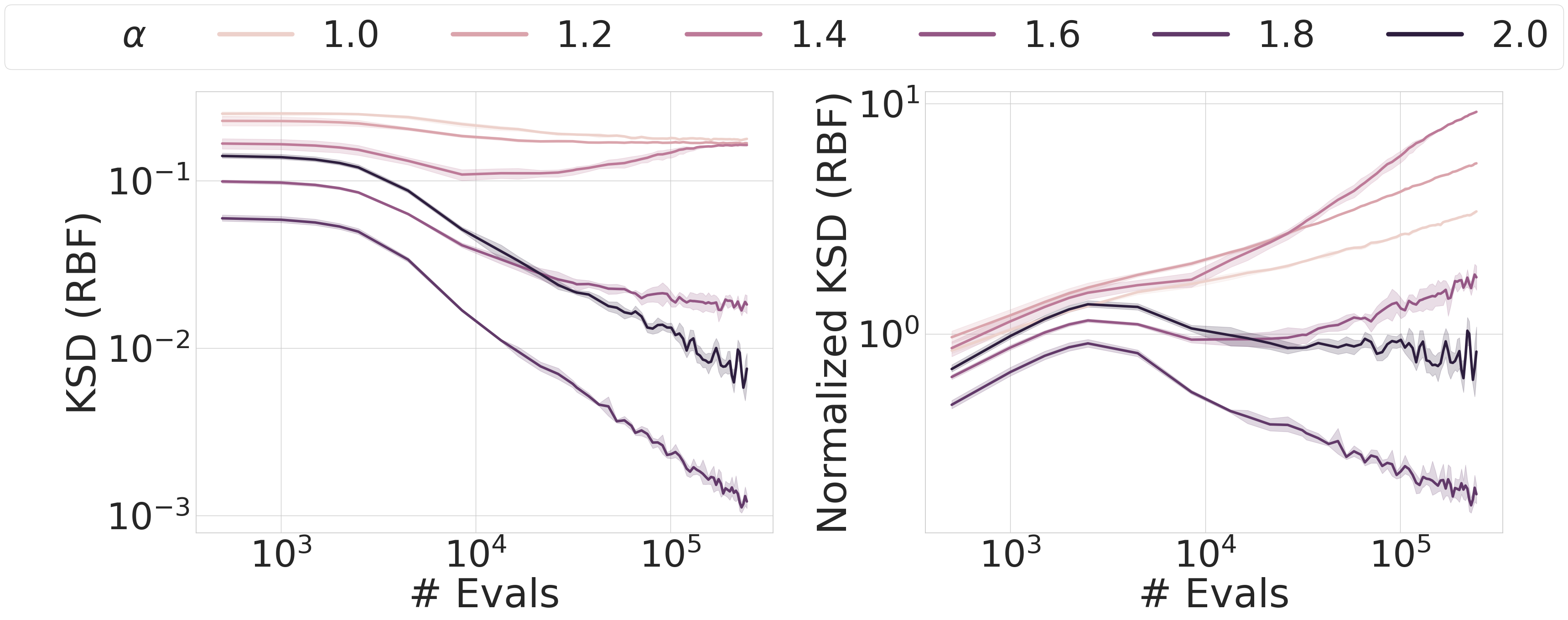}
    \caption{Sensitivity to dictionary growth rate parameter $\alpha$ on a bi-modal gaussian mixture problem using an RBF kernel. Lower KSD and lower Normalized KSD indicate more accurate representations of the target distribution. These results indicate that tuning the dictionary growth rate has potential to further improve our method.}
    \label{fig: alpha sensitivity rbf}
\end{figure}
In Figures \ref{fig: alpha sensitivity rbf} and \ref{fig: complexity adaptation rbf} we repeat the same experiments with the RBF kernel instead of the IMQ kernel. In Figure \ref{fig: alpha sensitivity} we vary the exponent $\alpha$ with budget $f(t)=\sqrt{t^\alpha\log t}/2$. When $\alpha=2.0$ we use linear growth rate $f(t)=t/2$. We observe the results are sensitive to parameter tuning on this toy problem, and that similar to the IMQ kernel experiment the best setting for both KSD and normalized KSD is $\alpha=1.8$. 

\subsection{Automatic Adaptation to Target Complexity}\label{sec: adaptation appendix}
\begin{figure}
    \centering
    \includegraphics[width=\textwidth]{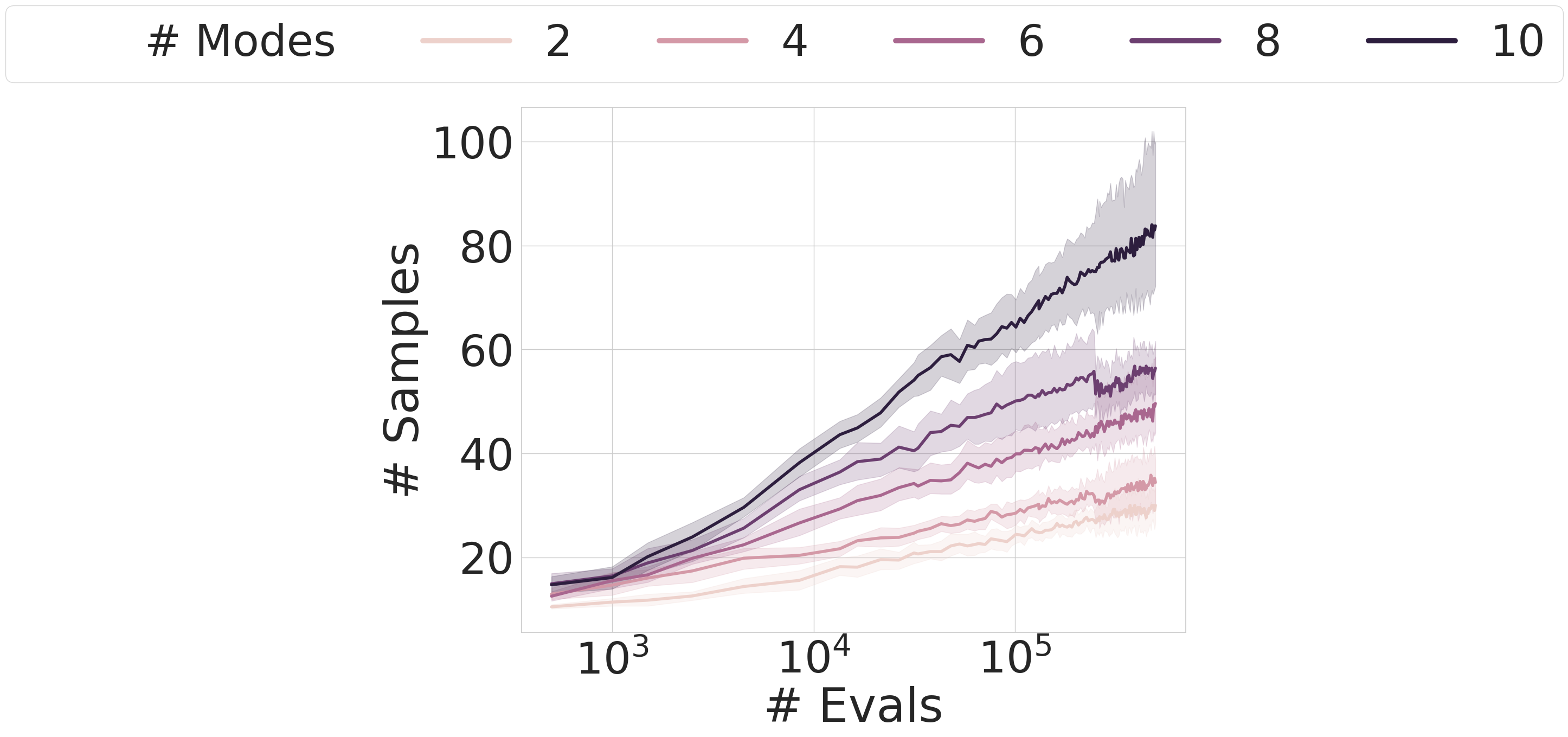}
    \caption{Base RBF kernel. As the number of modes of the Gaussian Mixture increases, our algorithm automatically adapts by increasing the number of samples.}
    \label{fig: complexity adaptation rbf}
\end{figure}
In Figure \ref{fig: complexity adaptation rbf} we observe that as the number of modes increases, so does the number of retained samples. This conclusion mirrors the IMQ kernel setting from Section \ref{sec: adaptivity}. 

%% file: sections/misc_appendix.tex
\section{Computational Complexity}\label{sec: computational complexity} 

This section addresses the per-step computational complexity of the problem 

The evaluation of \eqref{eq: least influential} does not require knowledge of the full symmetric kernel matrix $\mat{K}$ whose entries are $\mat{K}_{i,j}=k_0(\mat{z}_i,\mat{z_j})$ for points $\mat{z}_i\in\dict_{t-1}$. Assuming we maintain two vectors of size $|\dict_t|$ (row sums and per-sample KSD contributions) we require $\mathcal{O}\left(|\dict_{t-1}|\right)$ operations to evaluate \eqref{eq: least influential}. Evaluating the initial KSD to compute the thinning threshold $M$ requires summing the previous row sums with $\mathcal{O}\left(|\dict_{t-1}|\right)$ operations. Computing a new row requires $|\dict_{t-1}|$ evaluations of $k_0$. Updating previous row sums and KSD per-sample contributions requires $|\dict_{t-1}|$ additions. Thinning requires searching a vector of size $|\dict_{t-1}+1|$ for the maximum KSD contribution. A brute force search requires $\mathcal{O}\left(|\dict_{t-1}|\right)$ operations.
While thinning may occur more than once, we can amortize the cost of thinning across all time steps (since each sample can only be thinned once) to arrive at a worst case per-step complexity of $\mathcal{O}\left(t\right)$. In general, the thinning costs are orders of magnitude less expensive than the MCMC sampler costs.